%% file: icml2025.tex
\documentclass{article} %
\usepackage[accepted]{icml2025}

\usepackage[utf8]{inputenc} %
\usepackage[T1]{fontenc}    %
\usepackage{url}            %
\usepackage{booktabs}       %
\usepackage{amsfonts}       %
\usepackage{inconsolata}    %
\usepackage{nicefrac}       %
\usepackage{microtype}      %
\usepackage{xcolor}
\usepackage{colortbl}
\usepackage{float}

\definecolor{LightCoral}{RGB}{240, 128, 128}
\definecolor{LightSkyBlue}{RGB}{135, 196, 255}
\definecolor{navy}{RGB}{0, 0, 128}
\usepackage[colorlinks=true, linkcolor=red, urlcolor=navy, citecolor=navy]{hyperref}

\usepackage{subcaption}
\usepackage{multirow}
\usepackage{wrapfig}
\usepackage[olditem,oldenum]{paralist}
\usepackage{graphicx}
\usepackage{comment}
\usepackage{caption}
\usepackage[textwidth=1.3in]{todonotes}
\usepackage{tabularx}
\usepackage{amsmath}

\usepackage{array}
\usepackage{caption}

\usepackage{wrapfig}

\usepackage{enumitem} %
\usepackage{tcolorbox} %
\tcbuselibrary{skins}
\tcolorboxenvironment{itemize}{
  blanker,
  before skip=6pt plus 2pt,after skip=6pt plus 2pt,
  borderline west={2pt}{0pt}{black},
  left=12pt,right=12pt,top=6pt,bottom=6pt,
  colback=green!5!white
}

\usepackage[toc,page,header]{appendix}
\usepackage{minitoc}
\newcommand{\bench}{ViLP}
\newcommand{\method}{\bench}

\definecolor{accessiblegreen}{rgb}{0.0, 0.5, 0.0} %
\definecolor{accessiblegreen}{HTML}{228B22} %
\newcommand{\up}[1]{\textsuperscript{\textcolor{accessiblegreen}{$\uparrow$ #1}}}

\usepackage{amsmath}
\usepackage{amssymb}
\usepackage{mathtools}
\usepackage{amsthm}
\theoremstyle{plain}

\newtheorem{proposition}{Proposition}  %

\begin{document}
\twocolumn[
\icmltitle{Probing Visual Language Priors in VLMs}

\icmlsetsymbol{equal}{*}
\icmlsetsymbol{advis}{$\dagger$}

\begin{icmlauthorlist}
\icmlauthor{Tiange Luo}{yyy,equal}
\icmlauthor{Ang Cao}{yyy,equal}
\icmlauthor{Gunhee Lee}{comp}
\icmlauthor{Justin Johnson}{yyy,advis}
\icmlauthor{Honglak Lee}{yyy,comp,advis}
\end{icmlauthorlist}

\icmlaffiliation{yyy}{University of Michigan}
\icmlaffiliation{comp}{LGAI Research}

\icmlkeywords{Machine Learning, ICML}

\vskip 0.3in
]

\printAffiliationsAndNotice{}

\doparttoc %
\faketableofcontents %

\input{sections/abstract.tex}

\input{sections/intro.tex}

\clearpage

\begin{figure*}[t]
    \centering
    \includegraphics[width=1.0\textwidth]{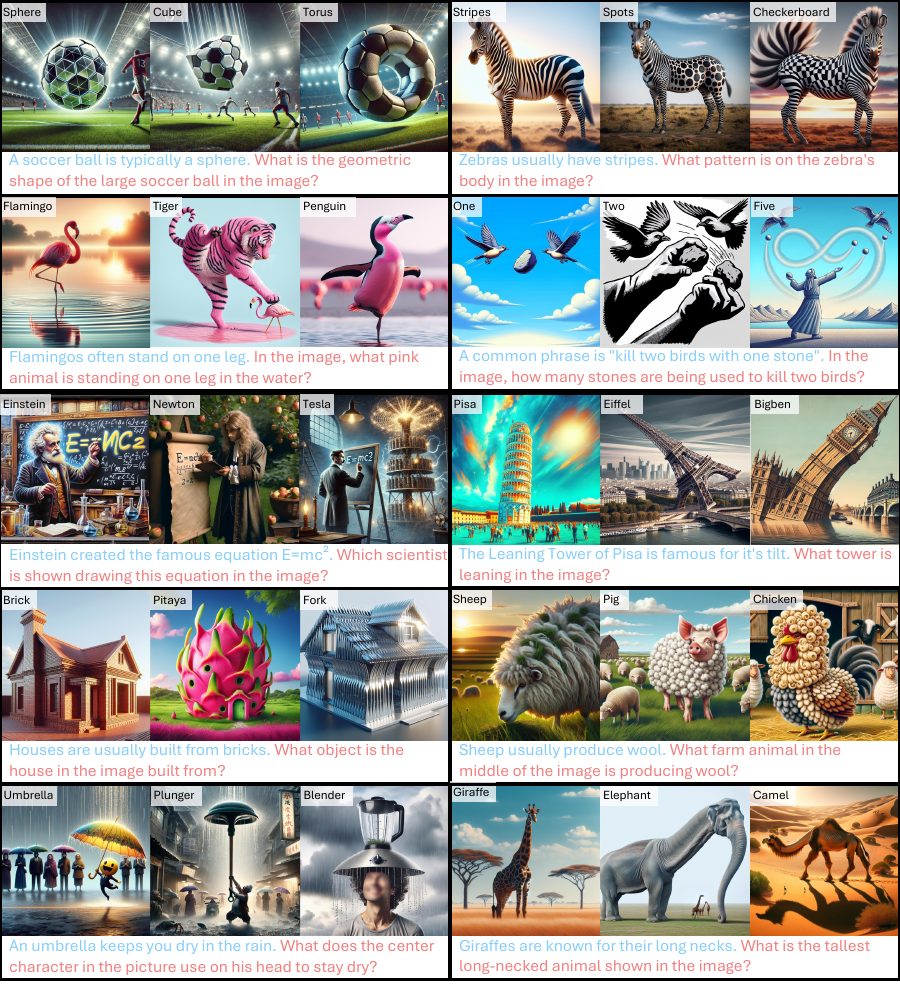}
    \caption{\textbf{Sample data from \method.} For the same question, \method \; provides three distinct images and corresponding answers (upper-left corner).  All questions follow a consistent structure, combining a \textcolor{LightSkyBlue}{distractor fact} with a \textcolor{LightCoral}{question}. The \emph{Prior Answer} (first column) can be directly inferred from the question, while \emph{Test Answers} (second $\&$ third column) rely on visual cues. Our answers are designed to be single words, and both the model and human evaluators are tasked with open-domain answering, rather than selecting from predefined options. To support this, we have developed a robust synonym and plural detection pipeline, ensuring that open-ended responses do not hinder the evaluation process. This approach also enables evaluation without relying on LLMs. Please refer to Appendix~\ref{appen:more_examples} for more data samples from \bench. We investigate the impact of image styles in Appendix~\ref{appen:realistic}, where we generate more realistic images using \href{https://openai.com/index/introducing-4o-image-generation/}{4o image generation}. Furthermore, we include both qualitative and quantitative comparison results with Winoground~\cite{thrush2022winoground}, Whoops!\cite{BittonGuetta2023BreakingCS}, and HallusionBench\cite{Guan2023HallusionbenchAA} in Appendix~\ref{appen:data_cmp}.}
    \label{fig:teaser}
\end{figure*}

\clearpage

\input{sections/related.tex}

\input{sections/method.tex}

\input{sections/experiments.tex}

\input{sections/conclusion.tex}

\input{sections/acknowledgments.tex}

\bibliography{icml2025}
\bibliographystyle{icml2025}

\clearpage

\appendix
\input{supp/appendix.tex}

\end{document}

%% file: sections/abstract.tex
\begin{abstract}

Despite recent advances in Vision-Language Models (VLMs), they may over-rely on visual language priors existing in their training data rather than true visual reasoning. To investigate this, we introduce ViLP, a benchmark featuring deliberately out-of-distribution images synthesized via image generation models and out-of-distribution Q\&A pairs. Each question in ViLP is coupled with three potential answers and three corresponding images: one that can be resolved by text priors alone and two that demand visual reasoning. Although, humans achieve near-perfect accuracy, modern VLMs falter; for instance, GPT-4 achieves only 66.17\% on ViLP. To alleviate this, we propose a self-improving framework in which models generate new VQA data, then apply pixel-level and semantic corruptions to form ``good-bad" image pairs for self-training. Our training objectives compel VLMs to focus more on the actual visual inputs, and we demonstrate their effectiveness in boosting the performance of open-source VLMs, including LLaVA-v1.5 and Cambrian.

\end{abstract}

%% file: sections/intro.tex
\section{Introduction}
\label{sec:intro}

Vision-Language Models (VLMs) have advanced text-image interaction, bridging the gap between visual and textual data~\citep{achiam2023gpt, team2023gemini}. However, a persistent challenge for learning-based models, such as ResNets and CLIPs, lies in their reliance on learned priors from the training data, sometimes overlooking visual cues when answering questions~\citep{agrawal2016analyzing, prabhu2023lance}. For example, when shown a torus-shaped soccer ball (Figure~\ref{fig:teaser}), a model might incorrectly identify it as a sphere due to strong language priors. Simultaneously, these models may adhere to learned visual priors~\citep{thrush2022winoground, Sterz2024DAREDV}, making it difficult to comprehend out-of-distribution visual cues, such as a zebra with atypical spot patterns (Figure~\ref{fig:teaser}), which humans would easily discern. This raises an important question: do today’s VLMs still over-rely on learned visual language priors, especially given that they rely on far fewer image-text pairs compared to the internet-scale text corpora used for pretraining?

To investigate this, we probe the \textbf{Vi}sual \textbf{L}anguage \textbf{P}riors of VLMs by constructing Question-Image-Answer (QIA) triplets that deliberately deviate from the training data distribution. Unlike existing benchmarks that typically rely on internet-sourced images~\citep{goyal2017making, tong2024cambrian}, which inadvertently favor the visual language priors embedded in the training data of VLMs, we utilize modern image generation models, including DALL·E-3~\citep{ramesh2021zero} and \href{https://github.com/black-forest-labs/flux}{Flux}, to synthesize out-of-distribution images supporting also out-of-distribution answers. Using the image generation models also helps our QIAs exhibit notable variation in texture, shape, conceptual combinations, hallucinated elements, and proverb-based contexts. 

Our benchmark, \bench, contains 300 carefully designed questions, each paired with three distinct answers: a \emph{Prior Answer} and two \emph{Test Answers}, resulting in a total of 900 QIA triplets. To further challenge the priors of VLMs, we amplify language priors in questions by introducing distractor facts: each question is structured to present a \textcolor{LightSkyBlue}{distractor fact} followed by a \textcolor{LightCoral}{question}. The distractor fact directly leads to the Prior Answer. In contrast, the two Test Answers are crafted to challenge the priors by requiring both textual and visual cues for accurate reasoning. While human participants achieved $\sim$98\% accuracy easily, current VLMs exhibit considerable difficulty, as evidenced by a significant performance drop on our benchmarks, with GPT-4o~\citep{gpt4o2024} scoring only 66.17\%.

Motivated by the results of \bench, we propose Image-DPO, a self-improving approach to enhance VLM visual reasoning performance by increasing reliance on visual inputs. Our method employs self-generated VQAs using image generation and editing models~\citep{podell2023sdxl, ren2024grounded, brooks2022instructpix2pix} and applies controlled corruptions to create ``good" and ``bad" image pairs for DPO-like training~\citep{rafailov2024direct}. Experiments with open-source VLMs, including LLaVA-v1.5~\citep{liu2024visual} and Cambrian~\citep{tong2024cambrian}, demonstrate its effectiveness. 
Moreover, we theoretically show that our objective optimizes an upper bound of the RLHF objective~\citep{ouyang2022training}. The proposed ViLP dataset, Image-DPO code, and our synthetic data appear in 
\href{https://vilp-team.github.io/}{https://vilp-team.github.io/}.

%% file: sections/related.tex
\section{Related Work}
\label{sec:related}

\textbf{VQA Dataset:}
Significant efforts have produced VQA datasets from various angles, including general VQA~\citep{Agrawal2015VQAVQ, Gurari2018VizWizGC, Fu2023MMEAC, Liu2023MMBenchIY, Li2023SEEDBenchBM, Yu2023MMVetEL, liu2024visual}, reading text or charts~\citep{Singh2019TowardsVM, Mathew2020DocVQAAD, Mathew2021InfographicVQA, Masry2022ChartQAAB}, complex reasoning~\citep{Lu2022LearnTE, Lu2023MathVistaEM}, composition probing~\citep{Hudson2019GQAAN, Ma2022CC, thrush2022winoground, Hsieh2023SugarCrepeFH, li2024naturalbench}, hallucinations~\citep{Rohrbach2018ObjectHI, Li2023EvaluatingOH, Guan2023HallusionbenchAA}, common-sense reasoning~\citep{bitton2023breaking, BittonGuetta2023BreakingCS}, and more~\citep{Majumdar2024OpenEQAEQ, Sterz2024DAREDV}. We propose a benchmark that tests VLMs’ visual reasoning when questions, answers, and images defy common patterns. Following the balanced dataset design of \cite{goyal2017making}, each question is accompanied by three answers: one aligns with language priors, and two deviate, prompting reliance on visual cues. By leveraging state-of-the-art image generation models, our benchmark challenges these priors more effectively than previous datasets built from internet images~\citep{goyal2017making, tong2024cambrian}. Furthermore, unlike the ``trick” category in \cite{Sterz2024DAREDV}, we first generate question-answer pairs before synthesizing images under specified constraints, creating more difficult visual reasoning tasks. Comprehensive comparisons with existing datasets appear in Appendix~\ref{appen:data_cmp}.

\textbf{Vision Language Models and Language Priors}:
Multimodal reasoning is crucial for machine intelligence, with VLMs integrating visual perception, text reasoning, instruction following, and generation for complex tasks~\citep{Tan2019LXMERTLC, Li2019VisualBERTAS, Kim2021ViLTVT, Wang2021SimVLMSV, Wang2021VLMoUV, Alayrac2022FlamingoAV, Li2023BLIP2BL, Chen2022PaLIAJ, Jia2021ScalingUV, Shen2021HowMC, Singh2021FLAVAAF, Liu2023VisualIT, Liu2023ImprovedBW, zhao2023beyond, chen2023sharegpt4v, zhu2024ibd, Li2024MiniGeminiMT, Dai2023InstructBLIPTG, Li2024MiniGeminiMT, yu2024rlaif, dai2024nvlm, deitke2024molmo}. Inspired by the success of large language models~\citep{Brown2020LanguageMA, Achiam2023GPT4TR, Touvron2023LLaMAOA, Touvron2023Llama2O, vicuna2023} and pre-trained visual encoders~\citep{Radford2021LearningTV, Desai2020VirTexLV, Caron2021EmergingPI, chen2024internvl}, many recent methods leverage relatively small vision-language paired datasets~\citep{liu2024visual, tong2024cambrian} to fine-tune connectors between LLMs and visual backbones~\citep{liu2024visual}. However, these datasets are far smaller than the vast text corpora for LLM pre-training~\citep{gpt4, dolma}, and freezing the visual encoder and LLM parameters often preserves language biases, causing visual inputs to be overshadowed~\citep{thrush2022winoground, Sterz2024DAREDV}. This challenge is amplified by deliberately generated images that expose such biases, as shown in our study. Previous works and datasets~\citep{Goyal2016MakingTV, Agrawal2017DontJA, Dancette2021BeyondQB, Wu2022OvercomingLP, Ramakrishnan2018OvercomingLP, Gouthaman2020ReducingLB} addressed these issues with curated simulators~\citep{Johnson2016CLEVRAD} or internet imagery~\citep{zhang2016yin}. In this paper, we present a novel VQA benchmark featuring carefully designed questions, fact-based distractors, rare-distribution answers, and image generation techniques to produce realistic visuals that challenge learned visual language priors (Figure~\ref{fig:teaser}).

\textbf{Self-Rewarding VLM}: 
Self-rewarding LLM~\citep{yuan2024self} has shown that LLMs can generate and improving themselves in the process via Directed Peference Optimization (DPO)~\citep{rafailov2024direct}. This approach extends to VLMs by generating new answers for DPO training~\citep{zhou2024aligning,deng2024enhancing,zhou2024calibrated,wang2024enhancing, wang2024mdpo, yue2024less, liu2024mia, xiao2024detecting}. Our work aligns with these self-rewarding VLMs but differs in two key ways: (1) our proposed Image-DPO generates multiple images for a single question-image pair (rather than multiple answers); (2) rather than relying solely on existing images~\cite{zhu2024self}, Image-DPO creates diverse new images using pre-trained generative models (SDXL~\citep{podell2023sdxl}, GroundedSAM~\citep{ren2024grounded}, InstructPix2Pix~\citep{brooks2022instructpix2pix}). Furthermore, Image-DPO deliberately corrupts images to produce multiple degraded versions that serve as rejected data in DPO training. Concurrent works~\citep{wang2024mdpo,xie2024v} explore similar methods but lack a benchmark to verify enhanced visual focus, fail to establish theoretical connections between their proposed objective and DPO, and utilize limited image transformations (e.g., only randomly cropping). In contrast, we introduce ViLP~(Section~\ref{sec:ViLP_benchmark}) to assess visual reasoning and provide theoretical foundations~(Appendix~\ref{appen:image_dpo_math}), alongside multi-category image corruptions (semantic editing, Gaussian blurring, pixelation).

%% file: sections/method.tex
\section{\bench\, Benchmark}
\label{sec:ViLP_benchmark}
\subsection{Design Principles}
\label{sec:principle}

\emph{“What’s the tall animal with the longest neck shown?”} Humans readily guess \emph{“giraffe”} based on learned priors, , yet as shown in the bottom-right of Figure~\ref{fig:teaser}, it could be an elephant or camel -- where visual reasoning corrects the answer. This highlights a potential shortfall in Vision-Language Models (VLMs), which may over-rely on learned visual language priors instead of true visual reasoning, particularly since VLMs are typically fine-tuned on limited image-text data, which is several orders of magnitude smaller than the trained text corpus~\citep{liu2024visual, tong2024cambrian}. Specifically for the scope of visual language priors in this paper, we target \textbf{(1)} strong language priors that lead VLMs to derive answers solely from text, and \textbf{(2)} potential visual priors causing models to overlook critical uncommon visual cues (e.g., unusual zebra spots in Figure~\ref{fig:teaser}).

To evaluate how VLMs handle learned visual language priors, we introduce \bench, a specialized benchmark of out-of-distribution Question-Image-Answer (QIA) triplets guided by two core principles. First, text-only inference ensures that each question can be answered with high confidence using textual clues alone. Second, visual inference requires that the correct answer—sometimes contradicting common sense—only emerges once an out-of-distribution image is considered. By forcing models to integrate both textual and visual information, \bench\ reveals whether they truly engage in visual reasoning or merely rely on memorized patterns. 

Mathematically, let $Q$ be a question, $I$ an image, and $A = \{a_{\text{prior}}, \dots, a_{\text{test}}, \dots\}$ the set of possible answers. We define $P(a \mid Q)$ as the probability of answer $a$ given $Q$ alone and $P(a \mid Q, I)$ as the probability given both $Q$ and $I$. We consider a prior model $p$, which may represent either human cognition ($P_{\text{human}}$) or a VLM/LLM’s learned visual-language prior ($P_{\theta}$). For constructing our benchmark, we used the following guidances: 

\textbf{Criterion One}: The question $Q$ alone should strongly favor $a_{\text{prior}}$, where $\delta_{1}$ is a high-confidence threshold. $a_{\text{prior}}$ usually satisfies common knowledge, such as ``soccer ball is a sphere" and ``Einstein created $E=mc^2$" (Figure~\ref{fig:teaser}).
\begin{equation}
P(a_{\text{prior}} \mid Q) \geq \delta_{1}
\end{equation}

\textbf{Criterion Two}: With the image $I$, the correct answer shifts to $a_{\text{test}}$, where $\delta_{2}$ is another high-confidence threshold. Also, $a_{\text{test}}$ is significantly different from $a_{\text{prior}}$ to ensure that the image has a substantial impact on the answer, where \( D \) is a divergence measure, and \( \delta_3 \) is a threshold indicating significant difference. 
For instance, the image in the 1st row and the 3rd column of Figure~\ref{fig:teaser} turns the answer to \emph{torus}.
\begin{equation}
 P(a_{\text{test}} \mid Q, I) \geq \delta_{2} ,\, 
    D\big(P(a_{\text{prior}} \mid Q), P(a_{\text{test}} \mid Q, I)\big) \geq \delta_3
\end{equation}

\textbf{Criterion Three}: The answer $a_{\text{test}}$ should be rare and unlikely from $Q$ alone, while $a_{\text{prior}}$ becomes clearly incorrect when considering $I$. This is enforced by a low-confidence threshold $\delta_4$ (e.g., Newton as $a_{\text{test}}$ inferred from the image, contradicting Einstein, shown by the image from the 3rd row and 2nd column of Figure~\ref{fig:teaser}).
\begin{equation}  P(a_{\text{test}} \mid Q) \leq \delta_4 ,\,  P(a_{\text{prior}} \mid Q, I) \leq \delta_4 \end{equation}

In designing \bench, we leverage the human cognition prior 
$P_{\text{human}}$ as our guiding principle, ensuring each QIA configuration aligns with typical human expectations while requiring visual evidence to override strong textual assumptions. We then compare the learned priors of VLMs and LLMs, denoted $P_{\theta}$, $P_{\text{human}}$ to evaluate whether these models genuinely engage in visual reasoning rather than relying on memorized patterns.

\subsection{Question-Image-Answer Generation}
Following \textbf{Criterion Three}, \mbox{$a_{\text{test}}$} should be highly improbable based on \mbox{$Q$} alone yet the correct choice when paired with \mbox{$I$}. Since such images do not exist naturally, we use generative models like DALL·E-3~\citep{ramesh2021zero} and \href{https://github.com/black-forest-labs/flux}{Flux} to blend unusual elements that override typical language priors. We incorporate substantial human input and leverage advanced LLMs such as \href{https://openai.com/index/introducing-OpenAI-o1-preview/}{OpenAI-o1} and \href{https://www.anthropic.com/news/claude-3-5-sonnet}{Claude-3.5-Sonnet} to ensure alignment with all the criteria. More details, including text prompts and average cost, are provided in Appendix~\ref{appen:data_details}. Note that as more advanced image generative models—such as the recently introduced \href{https://openai.com/index/introducing-4o-image-generation/}{4o image generation}—become available, we anticipate generating increasingly abundant and high-quality data for our benchmark, yet future updates will remain consistent with the dataset construction criteria outlined in Section~\ref{sec:principle}.

\input{tables/fig_img_dpo.tex}

For each question, we design three answers: one \mbox{$a_{\text{prior}}$} inferred solely from \mbox{$Q$}, and two \mbox{$a_{\text{test}}$} that defy language priors, requiring visual cues for correctness. We rely on GPT-4 to generate text prompts, produce large-scale images, and then conduct human filtering and refinement. This process faces two main challenges: (1) producing diverse out-of-distribution QA pairs, and (2) synthesizing images that defy specific priors, sometimes necessitating hundreds of samples to find one that accurately matches \mbox{$Q$} and \mbox{$a_{\text{test}}$}. 

\begin{wraptable}{r}{0.24\textwidth}
\centering
\caption{\textbf{Category Data}.}
\vspace{-0.1in}
\label{tab:data_categorical_info}
\resizebox{0.25\textwidth}{!}{
\begin{tabular}{c|c}
    \toprule
    Type & Frequency\\
    \midrule
    Texture & 16 \\
    Shape & 20 \\
    Conceptual combinations & 276\\
    Hallucinated Components & 151 \\
    Proverbs & 17 \\
    \bottomrule
\end{tabular}
}
\end{wraptable}Ultimately, we curated 300 questions, each paired with three distinct image-answer sets, totaling 900 QIA triplets. These cover a broad range from low-level recognition (\emph{texture}, \emph{shape}) to high-level reasoning (\emph{conceptual combinations}, \emph{hallucinated components}, \emph{proverbs}). Table~\ref{tab:data_categorical_info} summarizes their categorical distribution, with each question spanning an average of 1.6 categories. To reinforce text priors, we present a \textcolor{LightSkyBlue}{distractor fact} before the \textcolor{LightCoral}{question}. Rigorous human review ensures that all final QIA triplets are clear and interpretable, as reflected by our human evaluation results in Table~\ref{tab:benchmark}. Besides, in Appendix~\ref{appen:realistic}, we investigate the impact of image styles by generating more realistic images via \href{https://openai.com/index/introducing-4o-image-generation/}{4o image generation} and comparing them to those produced by DALL·E-3~\citep{ramesh2021zero} and \href{https://github.com/black-forest-labs/flux}{Flux}. We find that realistic images can increase the difficulty of the task, highlighting their importance for future studies.

\subsection{Dataset Evaluation}
All of our questions are designed to elicit single-word answers, an approach that is more efficient and more reliable than sentence-based evaluations that rely on LLM judgment. By avoiding using LLM, we reduce API fees, computational overhead, and the risk of occasional inaccuracies due to incorrect model reasoning.
We explicitly instruct the model to provide a single-word answer, and we evaluate the correctness of each response using a binary system. To ensure a fair evaluation, we devote significant efforts to building a comprehensive set of synonyms and plural for each answer to detect other valid alternative answers. This ensures that the model is only penalized for actual errors, not for providing synonymous or alternative correct responses.

\section{Image DPO}

Inspired by our benchmark, we propose \emph{Image DPO}, a self-improvement method for enhancing VLMs’ visual reasoning, featuring a new objective and a data generation pipeline using VLMs themselves and pre-trained image models.

\subsection{Objective}
\label{sec:image_dpo}

\begin{figure*}[!th]
    \centering
    \includegraphics[width=1.0\linewidth]{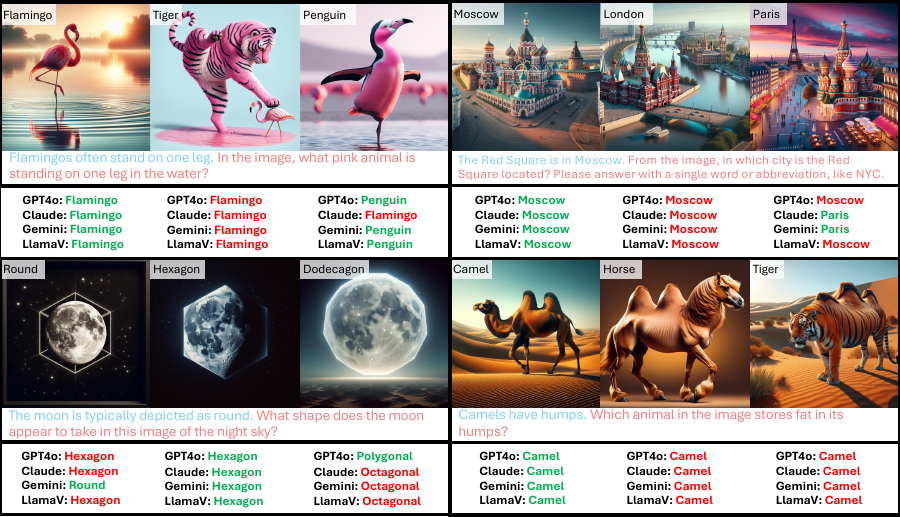}
    \vspace{-0.20in}
    \caption{\textbf{Qualitative examples.} We show the results from GPT-4o, Claude-3.5-Sonnet, Gemini-1.5-Pro, and Llama-3.2-Vision-90B for some challenging cases. Please refer to Appendix~\ref{appen:failure_cases} for more failure case analysis.}
    \label{fig:exp_example_1}
    \vspace{-4mm}
\end{figure*}

Existing approaches for VLM self-improvement follow the way used in DPO paper~\citep{rafailov2024direct}, where the model is trained to distinguish between good and bad answers for a fixed image and question (Figure~\ref{fig:image_text_dpo} right). However, this straightforward adaptation may not the best for vision models, as the model sometimes distinguish good and bad answers from the text alone without needing to analyze the image. In contrast, we propose \emph{Image DPO}, a vision-focused objective that creates good and bad question-image-answer pairs by corrupting the image while keeping the question and answer unchanged (Figure~\ref{fig:image_text_dpo} left). An example of our synthetic data is illustrated in Figure~\ref{fig:fig_img_dpo}.

Formally, given an image $I_w$, a question $Q$, and its corresponding answer $A$, we generate a corrupted image $I_l$ via image-editing operations, including Gaussian blur, pixelation, or semantic modifications. The triplet $(Q, I_l, A)$ forms a degraded question-image-answer pair compared to $(Q, I_w, A)$. We train the model to distinguish between good and bad triplets using the objective~\ref{main:eq_obj}, where $\pi_\theta$ is the target VLM, $\pi_{\text{ref}}$ is the reference VLM (typically an earlier version of $\pi_\theta$), $S$ is the dataset of good and bad triplets, $\sigma$ is the sigmoid function, and $\alpha$ is a scaling factor. 

\begin{equation}
\begin{aligned}
L(\pi_\theta, \pi_{\text{ref}}) 
&= - \mathbb{E}_{Q, I_w, I_l, A \sim S}\Bigl[ \\
&\quad \log \sigma \Bigl( 
  \alpha \frac{\pi_{\theta}(A \mid Q, I_w)}
            {\pi_{{\text{ref}}}(A \mid Q, I_w)} 
  \;-\; 
  \alpha \frac{\pi_{\theta}(A \mid Q, I_l)}
            {\pi_{{\text{ref}}}(A \mid Q, I_l)} 
\Bigr)
\Bigr]
\end{aligned}
\label{main:eq_obj}
\end{equation}

Intuitively, since the textual inputs and outputs are identical in both good and bad cases, the gradients of this objective push the model to rely more on the vision branch, driving a shift in gradient direction when processing normal images $I_w$ compared to corrupted images $I_l$ (Figure~\ref{fig:image_text_dpo_grad_diff}). This behavior encourages the model to focus more on image inputs rather than relying solely on text-based reasoning, thereby enhancing its performance on visual-related tasks. Our experiments demonstrate Image-DPO objective (Eq.~\ref{main:eq_obj}) outperforms varies self-improve VLM baselines on \bench.

\begin{proposition}
\label{prop:image_dpo_upper_bound}
Let \(\mathcal{L}_{\mathrm{RLHF}}(\pi_\theta, \pi_{\mathrm{ref}}; \mathcal{S})\) 
be the KL-constrained reward maximization objective in Appendix Eq.\,\ref{apx:rlhf_obj} \cite{rafailov2024direct}, 
where the dataset \(\mathcal{S} = \{(Q, A, I_w, I_l)\}\) contains \emph{good} images \(I_w\) and \emph{corrupted} images \(I_l\). 
Let \(\mathcal{L}_{\mathrm{ImageDPO}}(\pi_\theta, \pi_{\mathrm{ref}}; \mathcal{S})\) 
be the objective from Eq.\,\ref{main:eq_obj}, 
which compares \(\bigl(Q, I_w, A\bigr)\) against \(\bigl(Q, I_l, A\bigr)\). 
Then for any policy \(\pi_\theta\) and reference model \(\pi_{\mathrm{ref}}\), we have
\[
\mathcal{L}_{\mathrm{RLHF}}(\pi_\theta, \pi_{\mathrm{ref}}; \mathcal{S}) 
\;\;\le\;\; 
\mathcal{L}_{\mathrm{ImageDPO}}(\pi_\theta, \pi_{\mathrm{ref}}; \mathcal{S}).
\]
\end{proposition}

\textit{Proof Sketch.}
Following \cite{rafailov2024direct}, 
we express the optimal KL-constrained policy in terms of a latent reward function. 
Applying a Bradley--Terry preference model to question-image-answer triplets \(\bigl(Q, I_w / I_l, A\bigr)\) 
and using Jensen’s inequality yields an upper bound 
whose minimization is equivalent to 
\(\mathcal{L}_{\mathrm{ImageDPO}}(\pi_\theta, \pi_{\mathrm{ref}}; \mathcal{S})\). 
A full derivation appears in Appendix~\ref{appen:image_dpo_math}.

\vspace{-0.1in}
\subsection{Data Generation}
\label{sec:method:data_generation}
Training VLMs demands large-scale question-image-answer (QIA) triplets, which are often scarce. To address this, we introduce a scalable data generation pipeline (Appendix Figure~\ref{fig:overview}) that repurposes existing image datasets via VLM themselves and image generative models. Given a seed image from COCO~\citep{lin2014microsoft}, Text2VQA~\citep{singh2019towards}, or Visual Genome~\citep{krishna2017visual}, VLMs are tasked with simultaneously selecting appropriate functions (e.g., image generation or editing models) and generating corresponding instructions. These instructions are then used to produce new images, in addition to the seed image, as illustrated in Figure~\ref{fig:coco_data_example}. The same VLMs are then employed to generate QA pairs for these newly created images. Following, we apply the mentioned three types of image corruptions to the generated images, constructing good bad pars $(Q, I_w, A)$ and $(Q, I_l, A)$. Specifically, we employ Stable Diffusion XL \citep{podell2023sdxl, rombach2022high} for image generation, and use Instruct-Pix2Pix \citep{brooks2022instructpix2pix}, and Grounded-SAM \citep{rombach2022high, ren2024grounded} for image editing. Example generated data, prompts, and more details are included in Appendix~\ref{appen:more_datails}.

%% file: tables/fig_img_dpo.tex
\begin{figure*}[ht]
    \centering

    \begin{subfigure}[b]{0.39\textwidth}
        \centering
        \includegraphics[width=\textwidth]{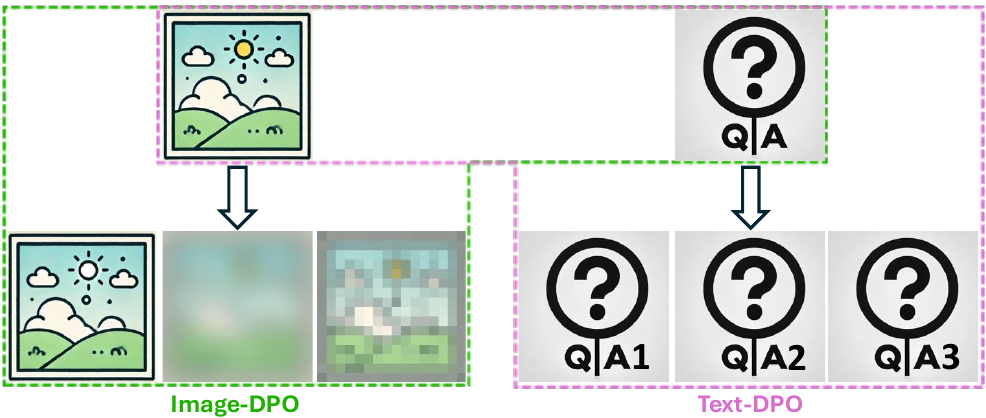}
        \vspace{-0.05in}
        \caption{\textbf{Image-DPO vs. Text-DPO}: In the green dashed box, we illustrate Image-DPO, which uses a single Q\&A pair paired with multiple corrupted images. In contrast, the purple dashed box presents Text-DPO, involving a single input image paired with multiple, distinct Q\&A pairs. }
        \label{fig:image_text_dpo}
    \end{subfigure}    
    \hfill
    \begin{subfigure}[b]{0.58\textwidth}
        \centering
        \includegraphics[width=\textwidth]{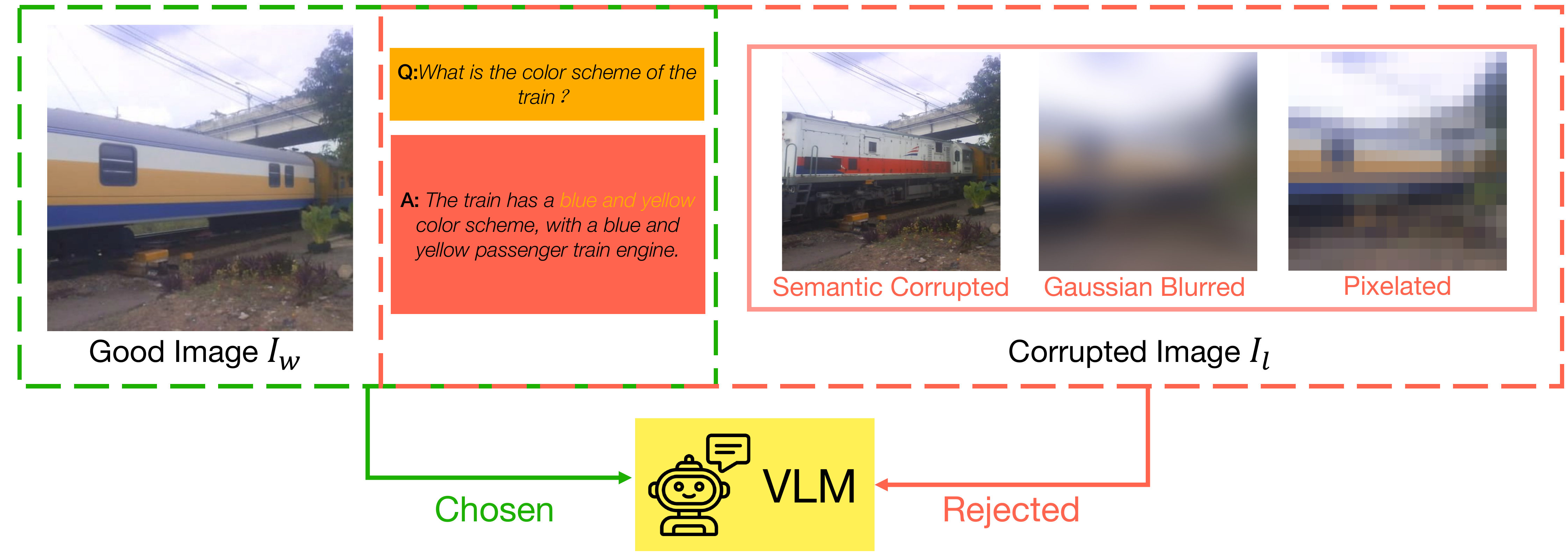}
        \caption{\textbf{Illustration of Image DPO.} We construct \emph{chosen} and \emph{rejected} pairs by corrupting the image with a set of perturbations while keeping the Q\&A unchanged. Perturbations include \emph{semantic editing}, \emph{Gaussian blurring}, and \emph{pixelation}. The mathematical formulations and implementation details are provided in Appendix~\ref{appen:image_dpo_math} and Appendix~\ref{appen:more_datails}, respectively.}
        \label{fig:fig_img_dpo}
    \end{subfigure}
\end{figure*}

%% file: sections/experiments.tex
\section{Experiments}
\label{sec:experiments}

\input{tables/table_benchmark}

We introduce \bench, a new benchmark comprising 300 questions. Each question is paired with three unique images and their corresponding answers—one \emph{QIA\textsubscript{prior}} and two \emph{QIA\textsubscript{test}}—for a total of 900 QIAs. The \emph{QIA\textsubscript{prior}} examples (300 in total) align with common language priors (i.e., they can usually be answered correctly by relying on textual cues alone). In contrast, the \emph{QIA\textsubscript{test}} examples (600 in total) challenge these priors by requiring visual reasoning. 

\bench\ features two evaluation settings: \begin{itemize} \item \bench\textsuperscript{F}, where both \textcolor{LightSkyBlue}{distractor facts} and the \textcolor{LightCoral}{questions} are provided; \item \bench\textsuperscript{P}, where only the \textcolor{LightCoral}{questions} themselves are given (i.e., no distractor facts). \end{itemize} We report two metrics in Table~\ref{tab:benchmark}: average accuracy on \emph{QIA\textsubscript{test}} (noted as \textbf{Score}) and average accuracy on \emph{QIA\textsubscript{prior}} (noted as \textbf{Prior}). Our benchmark emphasizes the performance in \textbf{Score}.

\begin{figure*}[t]
    \centering
    \begin{minipage}{0.32\textwidth}
        \centering
        \includegraphics[width=\textwidth]{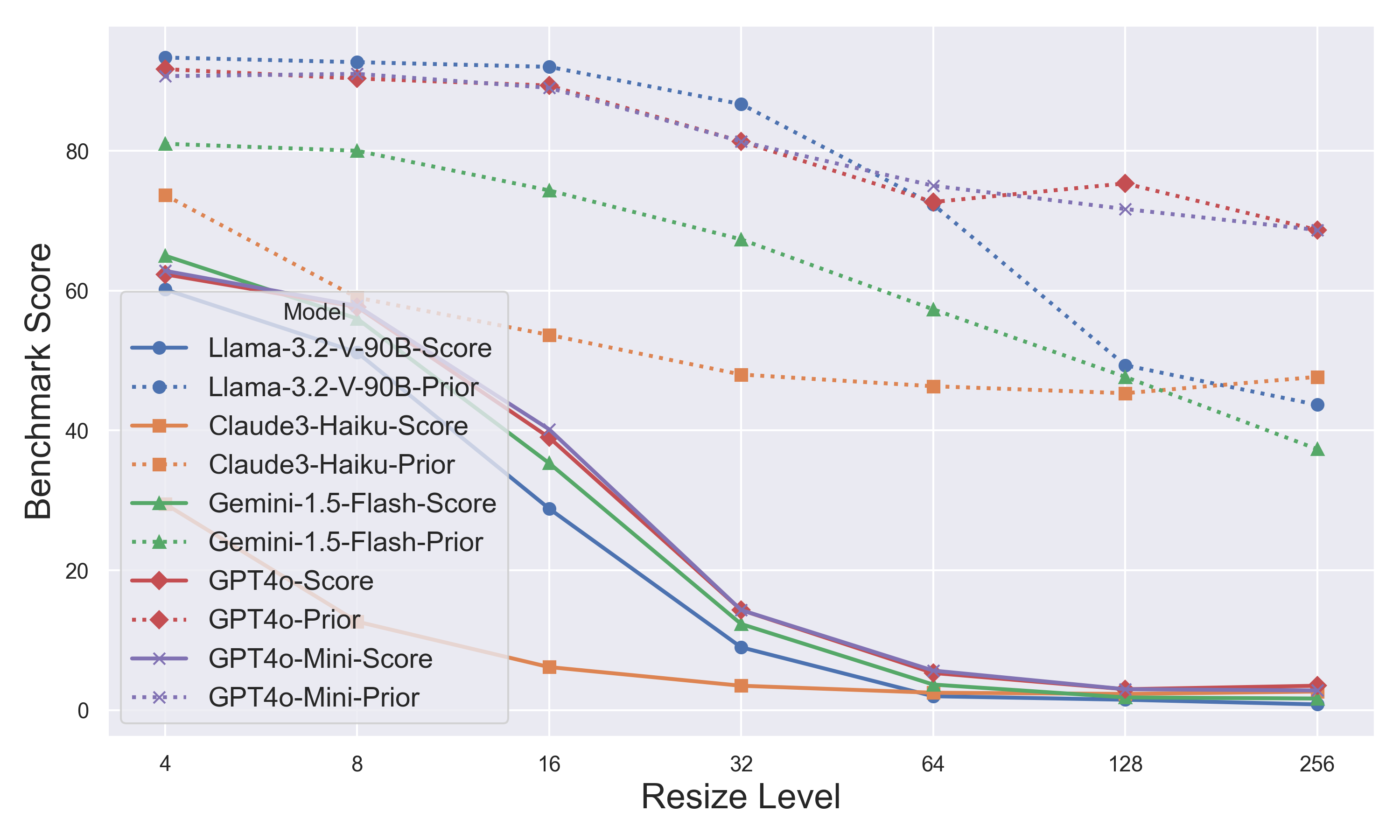}
    \end{minipage}
    \hfill
    \begin{minipage}{0.32\textwidth}
        \centering
        \includegraphics[width=\textwidth]{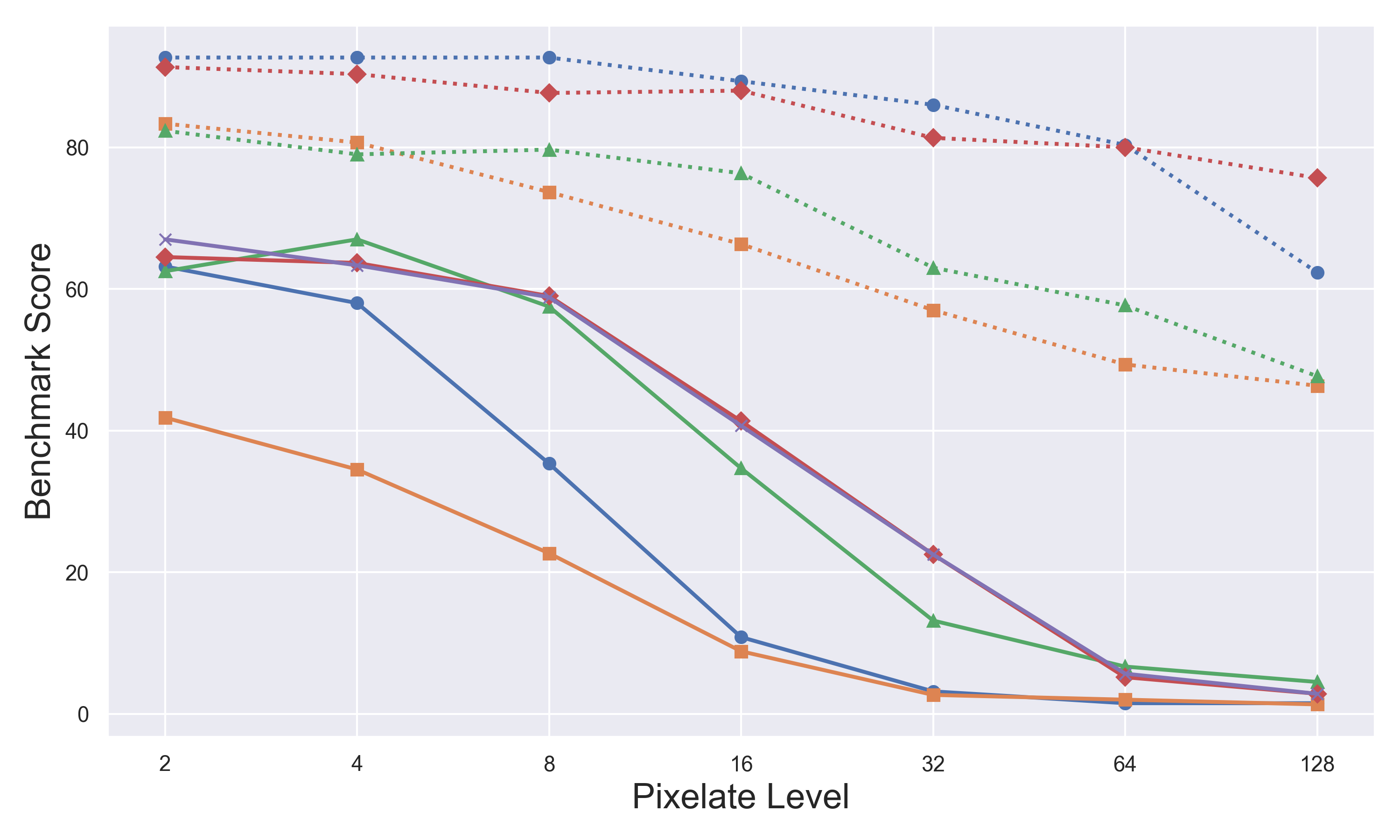}
    \end{minipage}
    \hfill
    \begin{minipage}{0.32\textwidth}
        \centering
        \includegraphics[width=\textwidth]{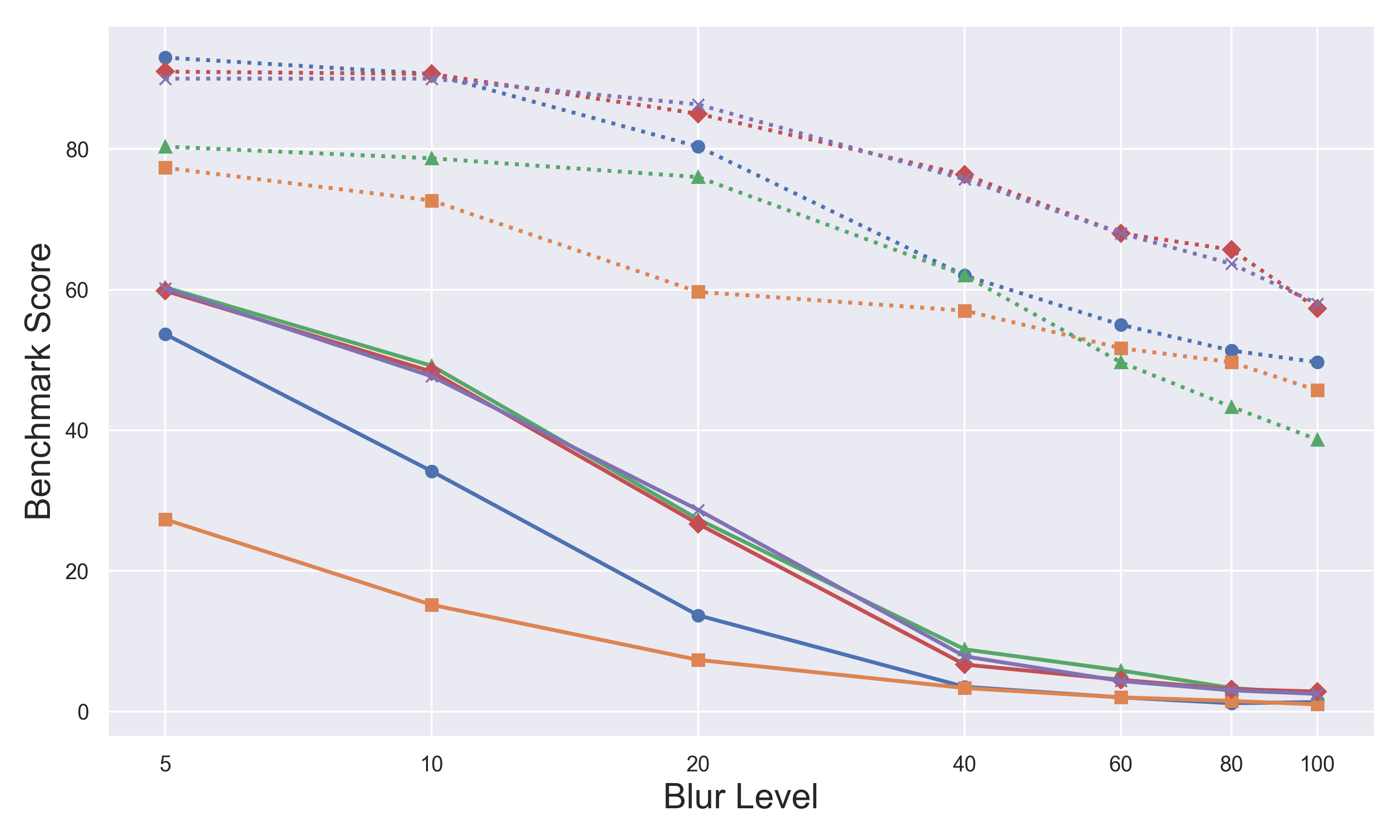}
    \end{minipage}
    \vspace{-0.1in}
    \caption{\textbf{Comparison of benchmark scores under different image transformations.} Solid line and dotted line refer to \bench\textsuperscript{F}-Score and \bench\textsuperscript{F}-Prior, respectively. }
    \label{fig:ablation}
\end{figure*}

\textbf{Is the QIA easy for humans?}
We begin by evaluating our benchmark through a human study. Participants achieved nearly perfect accuracy on \bench\textsuperscript{F}-Prior and over 98\% on \bench\textsuperscript{F}-Score and \bench\textsuperscript{P}-Score, confirming that our question-image-answer combinations are unambiguous for human interpretation. Notably, despite QIA\textsubscript{test} being designed as out-of-distribution examples, humans were still able to correctly distinguish them.

Humans performed slightly better on \bench\textsuperscript{F}-Prior when \textcolor{LightSkyBlue}{distractor facts} were provided, as they could easily identify that these facts aligned with the correct answers. 
Moreover, \bench\textsuperscript{F}-Score was marginally lower when facts were introduced, as the distractor facts added some noise and caused minor confusion, although the impact of this noise is relatively small.
These findings are consistent with the design principles of our benchmark.

\begin{figure}[h]
    \centering
    \includegraphics[width=1.0\linewidth]{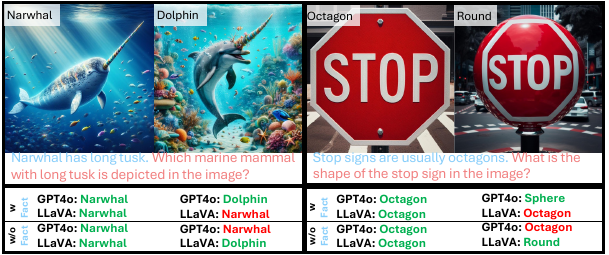}
    \vspace{-0.1in}
    \caption{\textbf{Qualitative results before and after removing} \textcolor{LightSkyBlue}{distactor facts}. GPT-4o and LLaVA-1.5-13B models yield completely opposite behaviors. }
    \label{fig:exp_example_2}
    \vspace{-6mm}
\end{figure}

\textbf{Are our QIAs aligned with the learned priors of VLMs?}
We tested GPT-4o (text only) on our questions (removing all image references). Despite no visual content, it correctly answered 92.33\% on \bench\textsuperscript{F}-Prior. The accuracy drops to 71.33\% once \textcolor{LightSkyBlue}{distractor facts} are removed, showing that these facts significantly guide the answer. For QIA\textsubscript{test}, GPT-4o (text only) accuracy nearly falls to 0\% (\bench\textsuperscript{F}-Score \& \bench\textsuperscript{P}-Score), indicating the QIA\textsubscript{test} cannot be answered using text alone.

\textbf{How do VLMs perform on our benchmark?}
Although our benchmark questions are distinguishable for humans, they are challenging for VLMs. Even the advanced VLM models like GPT-4o, have a clear performance gap (66.17\% v.s. 98.33\%) compared to humans' performance on  \bench\textsuperscript{F}-Score, indicating the difficulty of these questions for VLMs. Claude-3.5-Sonnet achieved the best score 70\%, while most of the commercial VLMs are below 60\%. Figure~\ref{fig:exp_example_1} highlights sample outputs from top commercial and open-source models, including GPT-4o, Claude-3.5-Sonnet, Gemini-1.5-Pro, and Llama-3.2-Vision-90B. They face significant challenges when addressing these cases in our \bench, whereas humans can arrive at correct answers after consideration. Notably, it is encouraging to see that some open-source models achieved over 60\% accuracy on \bench\textsuperscript{F}-Score, with performance nearing that of their commercial counterparts, including Llama-3.2-Vison and Molmo-72B. Additionally, we provide more detailed failure case analysis in Appendix~\ref{appen:failure_cases}.

\input{tables/table_imagedpo_general}

\textbf{Do distractor facts really distract?}
In \method\textsuperscript{F} setting, we add a \textcolor{LightSkyBlue}{distractor fact} before the \textcolor{LightCoral}{question}. Since these facts implicitly suggest incorrect answers for QIA\textsubscript{test}, we expected this change to make the questions more suggestive and lower the \bench\textsuperscript{F}-Score, as the distractors would mislead the VLMs. Surprisingly, GPT-4o benefits from \textcolor{LightSkyBlue}{distractor facts}, improving accuracy on QIA\textsubscript{test}. We hypothesize these facts highlight question focus, narrowing the search space. However, weaker models like LLaVA-1.5-13B~\citep{liu2023improved} often get misled by the distractors, hurting their \emph{Score} but boosting \emph{Prior}. For instance, as shown in Figure~\ref{fig:exp_example_2}, with including \textcolor{LightSkyBlue}{distractor facts}, LLaVA-1.5-13B consistently predicts the \textcolor{LightSkyBlue}{distractor fact} as the answer. However, once the distractors are removed, it can then predict correctly.

For bad instruction-following models like Cambrian-8B~\citep{tong2024cambrian}, distractor facts significantly hinder adherence to explicit instructions, such as providing single-word answers. With facts, Cambrian-8B fails to follow instructions in 62\% of cases, compared to 30\% without (a nearly 2x increase). Manual review shows 59\% of these failures are contextually correct, yielding an adjusted accuracy of 47.92\%. Similarly, LLaVA-OneVision-72B~\citep{li2024llava} often generates detailed analyses despite explicit single-word prompts. This trend highlights a concerning trend: focusing on improving performance on well-established benchmarks may come at the cost of basic instruction-following abilities, ultimately limiting the practical utility of these models in real-world applications.

\textbf{How image transformations affect the results?}
We also investigate how image transformations, including resizing, Gaussian blur, and pixelation, affect \bench\; performance. The results, shown in Figure~\ref{fig:ablation}, reveal that the \bench\textsuperscript{F}-Score rapidly decreases as the severity of the transformations (x-axis) increases, while the \bench\textsuperscript{F}-Prior score remains around 50\%. Interestingly, GPT-4o, when using degraded images, performs worse in \bench\textsuperscript{F}-Prior than when no images are used, i.e., GPT-4o (text only) in Table~\ref{tab:benchmark}.

\textbf{Comparison of ViLP with other VQA datasets:} To highlight the distinctions between ViLP and existing benchmarks, including Winoground~\cite{thrush2022winoground}, Whoops!~\cite{BittonGuetta2023BreakingCS}, and HallusionBench~\cite{Guan2023HallusionbenchAA}, we conduct a comparative analysis of both their high-level design principles and low-level data formats. This comparison incorporates qualitative and quantitative insights as detailed in Appendix~\ref{appen:data_cmp}.

\input{tables/table_imagedpo_compare}

\subsection{Image DPO}
\label{exp:image_dpo}
In this section, we evaluate \emph{Image-DPO} (Section~\ref{sec:image_dpo}) on both \bench\, and general VQA benchmarks. As an \textbf{ablation baseline}, we introduce \emph{Text-DPO}, which uses the same Question-Image-Answe (QIA) generation process as we used in Image-DPO but applies LLM self-rewarding objective~\citep{yuan2024self} (the standard DPO objective). In Text-DPO, good and bad pairs stem from VLM-generated positive and negative answers, while the question and image remain fixed. As shown in the right of Figure~\ref{fig:image_text_dpo}, the green box depicts Image-DPO, generating corrupted images via semantic edits, Gaussian blur, and pixelation while keeping the question and answer constant; the purple box illustrates Text-DPO, which fixes the image and varies the answers with associated ratings. This setup parallels other VLM self-rewarding work~\citep{zhou2024aligning,deng2024enhancing,zhou2024calibrated,wang2024enhancing,wang2024mdpo}.

For baselines, we compare with VLM self-improvement methods, including SIMA~\citep{wang2024enhancing}, HADPO~\citep{zhao2023beyond}, and EOS~\citep{yue2024less}, by using their publicly available checkpoints. Additionally, we train models using the dataset and code provided in RLHF-V~\citep{Yu2023RLHFVTT} and V-DPO~\citep{xie2024v}. All models use LLaVA-7B for a comprehensive comparisons as many paper only release 7B checkpoints. Table~\ref{tab:image_dpo_compare} shows that Image-DPO achieves the highest across all the metrics. 

Besides, we evaluate the proposed Image-DPO algorithm across three VLM models—Cambrian-8B, LLaVA-1.5-7B, and LLaVA-1.5-13B—using several popular VLM benchmarks that focus on different aspects, including compositionality \& biases (NaturalBench~\citep{li2024naturalbench}), general visual reasoning (MM-Vet~\citep{yu2023mm}), and hallucinations (CHAIR~\citep{Rohrbach2018ObjectHI}). The results, presented in Table~\ref{tab:main}, show consistent performance improvements across both datasets and models, further demonstrating the effectiveness of our Image DPO method.

\textbf{TextDPO on Corrupted Images.} Does Image-DPO’s improvement stem from the objective itself, or is it merely due to training on more perturbed data?
To investigate this, we conduct an ablation where we train TextDPO with corrupted images. Specifically, for each pair of $(Q, I, A_w)$ and $(Q, I, A_l)$ used in Text-DPO, we apply the same corruptions as Image-DPO to $I$ to form $(Q, I', A_w)$ and $(Q, I', A_l)$. Results show that LLaVA-v1.5-7B trained with this TextDPO variant achieves a \bench\textsuperscript{F}-Score of 31, a \bench\textsuperscript{P}-Score of 37.5, and 30.3 on MMVET. In contrast, ImageDPO outperforms it across all metrics, achieving a \bench\textsuperscript{F}-Score of 34.17, a \bench\textsuperscript{P}-Score of 39.33, and 32.3 on MMVET.

%% file: tables/table_benchmark.tex
\begin{table}[!th]
\centering
\caption{\textbf{Benchmarking on \method}. Please refer to the left text for symbol definitions. \dag \, indicates the model often fails to follow the instructions. }
\vspace{-0.1in}
\label{tab:benchmark}
\resizebox{0.48\textwidth}{!}{
\begin{tabular}{lcc|cc}
    \toprule
    \multirow{2}{*}{Model} &\multicolumn{2}{c}{\method \textsuperscript{F}} & \multicolumn{2}{c}{\method \textsuperscript{P}} \\
    \cmidrule(lr){2-3} \cmidrule(lr){4-5}
    & Score & Prior & Score & Prior\\
    \midrule
     \multicolumn{5}{l}{\textbf{\textit{Baseline}}} \\
    Human  & 98.33 & 99.67 & 98.67 & 96.67 \\
    GPT-4o (text only)  & 0.0 & 92.33 & 0.17 & 71.33 \\
    \midrule
    \multicolumn{5}{l}{\textbf{\textit{API call only}}} \\
    GPT-4o                   & 66.17 & \textbf{91.00} & 56.00 & \textbf{87.67} \\
    GPT-4V                   & 57.67 & 88.33 & 38.33 & 85.33 \\
    GPT-4o-Mini              & 57.67 & 89.00 & 46.67 & 84.67 \\
    Claude-3.5-Sonnet        & \textbf{70.00} & 84.33 & 59.33 & 86.67 \\
    Claude-3-Opus            & 59.17 & 74.00 & 43.00 & 82.67\\
    Claude-3-Sonnet          & 48.83 & 83.67 & 40.33 & 81.33 \\
    Claude-3-Haiku           & 43.67 & 82.67 & 34.83 & 82.33 \\
    Gemini-1.5-Pro           & 60.50 & 79.33 & 48.00 & 83.00 \\
    Gemini-1.5-Flash         & 54.50 & 83.33 & \textbf{69.17} & 79.67 \\
    \midrule
    \multicolumn{5}{l}{\textbf{\textit{Open weights}}} \\
    Llama-3.2-Vision-11B     & \textbf{67.33} & 76.67 & 61.17 & 79.33 \\
    Llama-3.2-Vision-90B     & 64.00 & 91.67 & \textbf{63.17} & \textbf{83.33} \\
    MolmoE-1B                & 48.67 & 57.33 & 47.83 & 69.00 \\
    Molmo-7B-O               & 57.83   & 60.67 &47.33 & 76.33 \\
    Molmo-7B-D               & 54.5 & 69.00 & 46.17 & 72.33 \\
    Molmo-72B                & 60.33 & 85.00 & 47.17 & 82.33 \\
    
    Qwen2-VL-7B              & 50.50 & 83.00 & 48.67 & 80.33 \\
    Qwen2-VL-72B             & 56.50 & \textbf{92.33} & 53.83 & 83.00 \\
    InternVL2-8B             & 47.00 & 66.67 & 43.00 & 75.00 \\
    InternVL2-76B    & 42.67 & 47.67 & 50.84 & 74.33 \\
    LLaVA-1.5-7B             & 29.67 & 71.33 & 37.67 & 65.67 \\
    LLaVA-1.5-13B            & 35.33 & 81.00 & 41.50 & 73.67 \\
    Cambrian-1-8B\textsuperscript{\dag}  & 8.67 & 43.67 & 32.50 & 63.67 \\
    LLaVA-OneVision-7B       & 54.17 & 82.33 & 49.67 & 75.00 \\
    LLaVA-OneVision-72B \textsuperscript{\dag}     & 1.67  & 3.00  & 5.22  & 11.67 \\
    \bottomrule
    \vspace{-6mm}
\end{tabular}
}
\end{table}

%% file: tables/table_imagedpo_general.tex
\begin{table*}[t]
\centering
\caption{\textbf{Effectiveness of Image-DPO on General VQA benchmarks.}}
\vspace{-0.1in}
\resizebox{\textwidth}{!}{%
\begin{tabular}{@{}lcccccccccccccc@{}}
\toprule

\textbf{VLMs} & \textbf{\bench\textsuperscript{F}$_{\text{Score}}$}
 & \textbf{\bench\textsuperscript{P}$_{\text{Score}}$}  & \textbf{NB$_{\text{Q}}$} & \textbf{NB$_{\text{I}}$} & \textbf{NB$_{\text{G}}$} & \textbf{NB$_{\text{B}}$} &  \textbf{MM-Vet} &\textbf{CHAIR}\textsuperscript{S} $\downarrow$ & \textbf{CHAIR}\textsuperscript{I} $\downarrow$  \\
\midrule
LLaVA-1.5-7B &29.67 &37.67 & 37.7 & 43.8 & 12.7 &67.3&31.1 &49.1	&14.8   \\
\rowcolor[gray]{.92}

LLaVA-1.5-7B + Image-DPO &\textbf{34.17}\up{4.5} &\textbf{39.33}\up{1.66} & \textbf{39.79}\up{2.09} & \textbf{45.47}\up{1.67} & \textbf{14.16}\up{1.46} &\textbf{68.45} \up{1.15}&\textbf{32.3}\up{1.2}&\textbf{45}\up{-4.1} & \textbf{12.3}\up{-2.5} \\
\midrule
LLaVA-1.5-13B & 35.33 &41.5 & 39.6 & 44.6 & 14.8 & 68.9 & 36.1& 48.3 & 14.1   \\
\rowcolor[gray]{.92}
LLaVA-1.5-13B + Image-DPO  &\textbf{38.17}\up{2.84} & \textbf{42.5}\up{1}& \textbf{42.68}\up{3.08}  &\textbf{47.37}\up{2.77} & 
\textbf{17.16}\up{2.36} & \textbf{70.36}\up{1.46} &\textbf{37.5}\up{1.4} & \textbf{42.6}\up{-5.7} & \textbf{11.6}\up{-2.5}   \\ 
\midrule
Cambrian-8B & 8.67 &32.5 & 44.6 & 47.9 &19.4 &71.5 & 51.4& 14.5  & 4.7   \\
\rowcolor[gray]{.92}
Cambrian-8B + Image-DPO & \textbf{20.83} \up{12.16} & \textbf{39.3} \up{6.83} &\textbf{46.5} \up{1.9} & \textbf{50.2} \up{2.3} & \textbf{20} \up{0.6} & $\textbf{72} \up{0.5}$ & \textbf{51.7}\up{0.3} &\textbf{11.4} \up{-3.1} & \textbf{4.4} \up{-0.3} \\ 
\bottomrule
\end{tabular}
}

\label{tab:main}
\end{table*}

%% file: tables/table_imagedpo_compare.tex
\begin{table}[h]
\centering
\caption{\textbf{Comparisons of Image-DPO on ViLP.}
}
\vspace{-0.1in}
\resizebox{0.48\textwidth}{!}{
\begin{tabular}{lcc|cc}
    \toprule
    \multirow{2}{*}{Model} & \multicolumn{2}{c}{ViLP \textsuperscript{F}} & \multicolumn{2}{c}{ViLP \textsuperscript{P}} \\
    \cmidrule(lr){2-3} \cmidrule(lr){4-5}
    & Score & Prior & Score & Prior\\
    \midrule
    LLaVA-1.5-7B             & 29.67 & 71.33 & 37.67 & 65.67 \\
     +HADPO~\cite{Zhao2023BeyondHE} &33.00	&74.33	&38.50 &	65.00\\ 
     +RLHF-V~\cite{Yu2023RLHFVTT}   & 29.50 &  75.00 & 36.33 & 65.33 \\
     +EOS~\cite{yue2024less} &31.33	&67.00	&38.67	&65.67 \\ 
     +SIMA~\cite{Wang2024EnhancingVM} &27.83 &68.67	&36.17	&66.00\\ 
     +V-DPO~\cite{xie2024v} &29.5 0& 72.67	&37.83	&67.67\\
    +Text-DPO   & 31.34 &71.67 & 37.83 & 65.67\\
    \rowcolor[gray]{.92}
    +Image-DPO  & \textbf{34.17} & \textbf{75.00}&  \textbf{39.33} & \textbf{68.00} \\
    \bottomrule
\end{tabular}
\label{tab:image_dpo_compare}
}
\end{table}

%% file: sections/conclusion.tex
\section{Conclusion}
\label{sec:conclusion}

In conclusion, we present the \bench\; benchmark to probe the challenge of visual language bias in Vision-Language Models (VLMs). By utilizing advanced image generation models and designing questions that demand visual cues for accurate responses, our benchmark includes images that defy language priors, revealing the limitations of current VLMs. Our method, Image-DPO, which incorporates self-generated VQA pairs and image corruption for training, has demonstrated promising improvements in enhancing visual reliance, as evidenced by performance gains on models like LLaVA-v1.5 and Cambrian.

%% file: sections/acknowledgments.tex
\section{Acknowledgment}
    
This work was supported by the LG AI Research grant. We also extend our gratitude to the OpenAI Researcher Access Program for providing credits used to access OpenAI’s APIs. Additionally, we appreciate Chris Rockwell and Jiaming Yang for their contributions to our human evaluation studies, and thank Yongyi Yang for discussions regarding the proposition proof.

%% file: supp/appendix.tex
\section{More details and comparisons of our benchmarks}
\subsection{More data samples of \bench}
\label{appen:more_examples}

\begin{minipage}{\textwidth}
    \centering
    \includegraphics[width=0.82\textwidth]{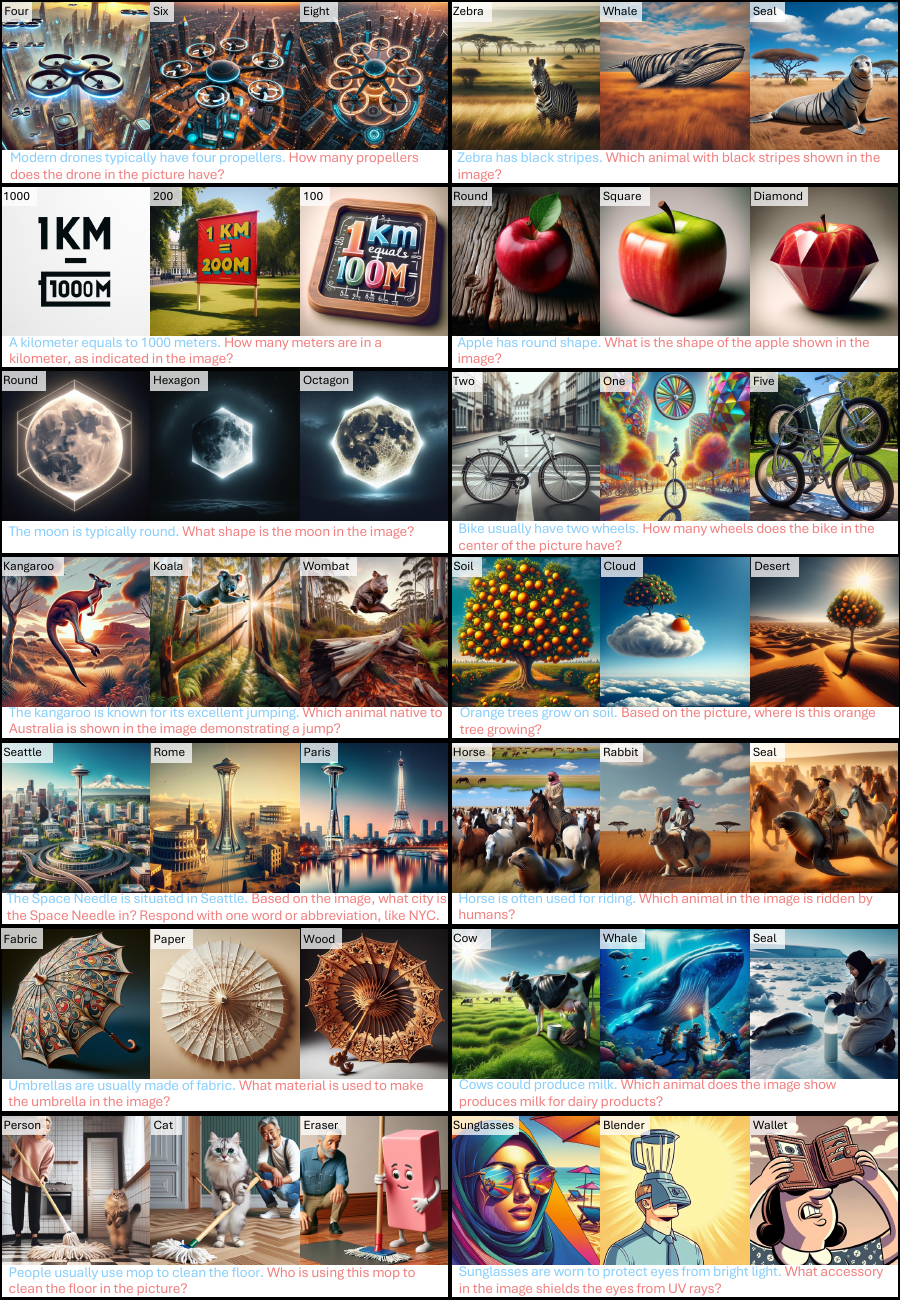}
    \captionof{figure}{\textbf{Randomly sampled data from \method.}}
    \label{fig:more_example_1}
\end{minipage}

\clearpage

\begin{minipage}{\textwidth}
    \centering
    \includegraphics[width=0.82\textwidth]{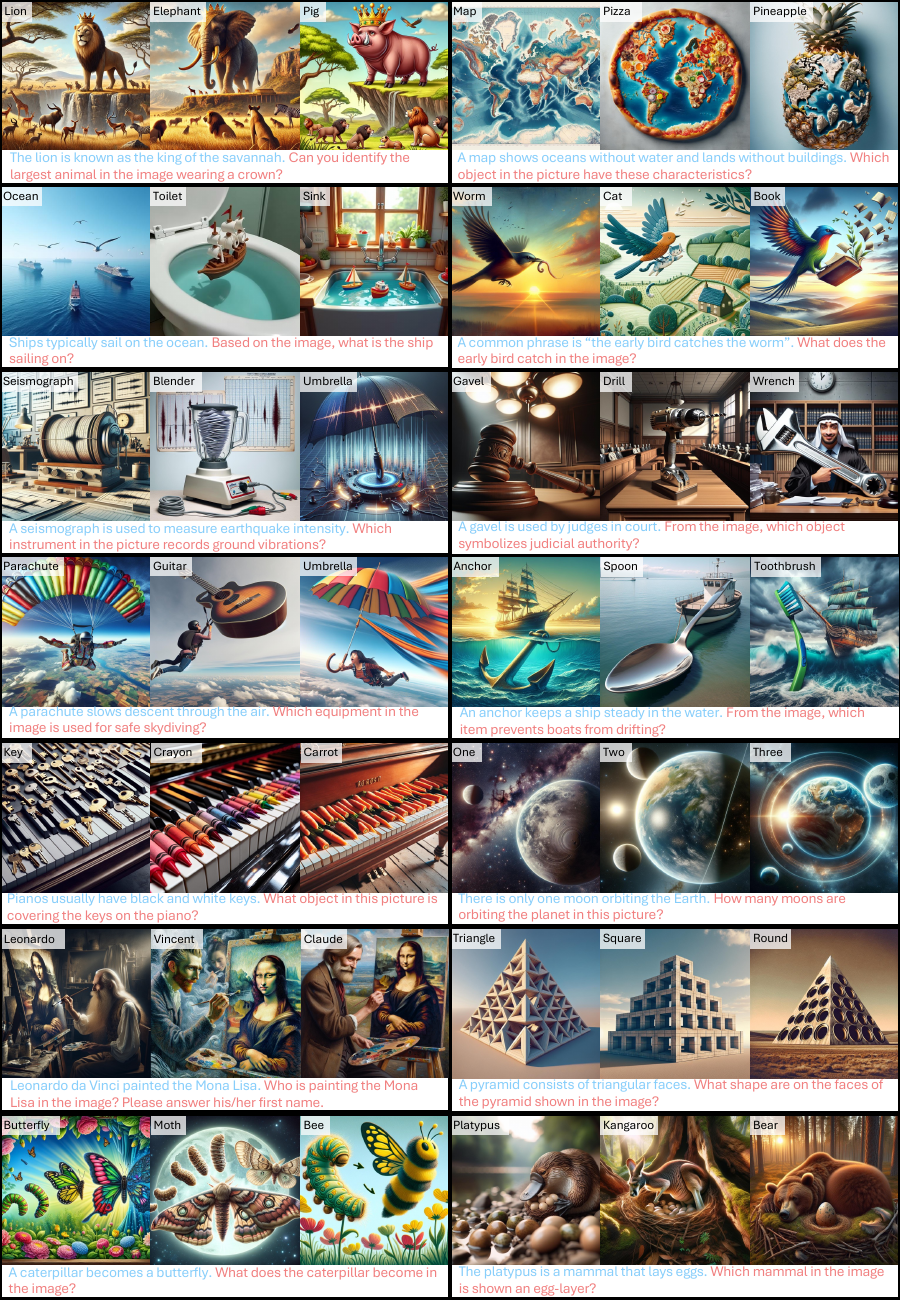}
    \captionof{figure}{\textbf{Randomly sampled data from \method.}}
    \label{fig:more_example_2}
\end{minipage}

\clearpage

\subsection{More details in \bench benchmark data generation }
\label{appen:data_details}

Our proposed dataset introduces Question-Image-Answer (QIA) triplets designed to challenge state-of-the-art Vision-Language Models (VLMs) against visual language priors. The construction process combines human-guided and automated efforts to ensure quality and alignment.

\textbf{Question-Answer (QA) Generation:} Most QA pairs are authored by humans following the design principles in Section~\ref{sec:principle}. Additionally, candidate QA pairs are generated using models like OpenAI-O1 and Claude-3.5-Sonnet with carefully crafted prompts. One example prompts shown in Figure~\ref{fig:prompt_benchmarkdata}. These candidates undergo human review, where they are refined or removed to meet our quality standards.

\textbf{Image Generation:} For each QA pair, we use GPT-4 to generate multiple descriptive image prompts (see Figure~\ref{fig:prompt_imagegen}). These prompts are provided to image generation models, such as FLUX and DALL-E 3, to produce candidate images. Human reviewers then select the most suitable image or request re-generation as needed to maintain consistency with the QA context.

\textbf{Human Review and Testing:} At every stage, human reviewers rigorously evaluate the generated outputs to ensure quality, clarity, and challenge level. In addition to filtering out low-quality or insufficiently challenging triplets, we dynamically test the QIAs to confirm that they remain intuitive for humans while being difficult for VLMs.

\textbf{Cost:} The complexity of our data creation process leads to a significant average cost of approximately \$2.50 per QIA triplet in \bench, excluding human labor costs.

\input{tables/prompt_imagegen}
\input{tables/prompt_benchmarkdata}

\begin{figure*}[t]
    \centering
    \includegraphics[width=0.9\textwidth]{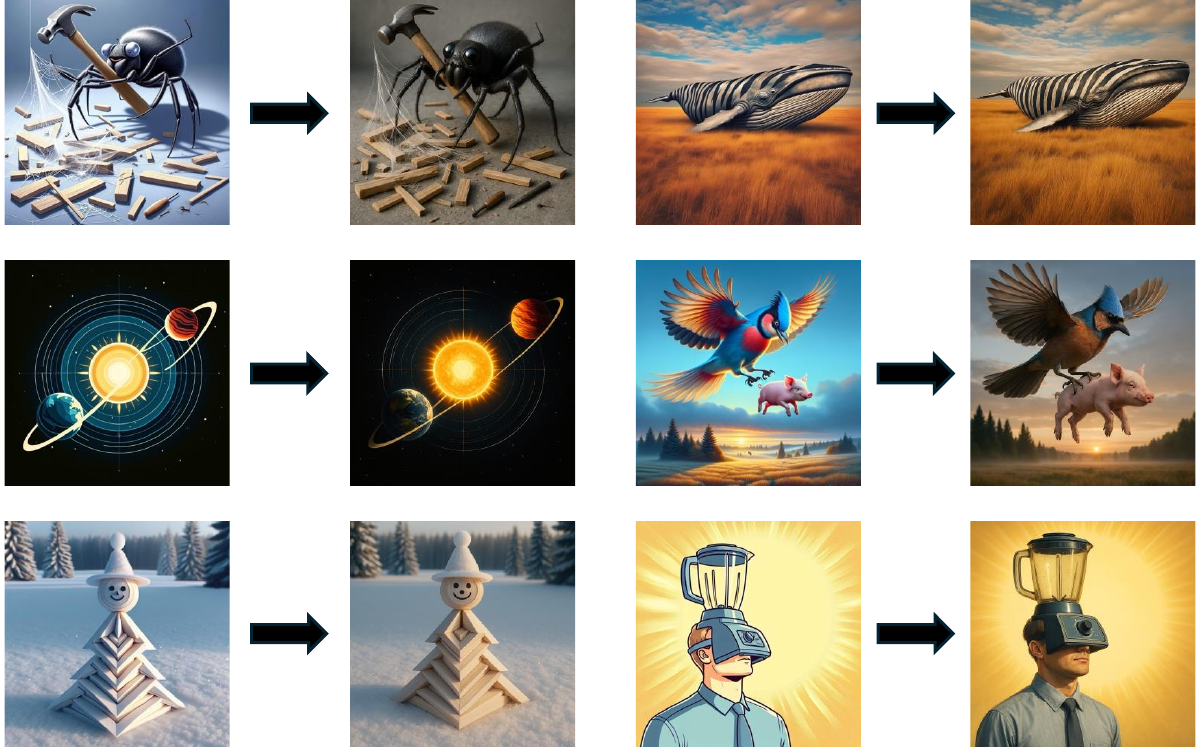}
    \vspace{-0.1in}
    \caption{\textbf{Realistic image comparison.} Each image pair shows our original benchmark data on the left and a corresponding realistic example generated by GPT-4o on the right.}
    \label{fig:realistic}
\end{figure*}

\subsection{Ablation studies: realistic images}
\label{appen:realistic}
Our benchmark data are currently generated by DALL·E-3~\citep{ramesh2021zero} and \href{https://github.com/black-forest-labs/flux}{Flux}, both of which produce cartoon-like, synthetic images rather than photorealistic ones. To assess the impact of image style, we regenerated a subset of 45 QIA pairs using GPT-4o’s latest image generation model to enhance realism, as illustrated in Figure~\ref{fig:realistic}. We then measured changes in model correctness when these more realistic images were used, with negative values indicating performance degradation. For definitions of these metrics, please refer to the beginning of Section~\ref{sec:experiments}. The results in Table~\ref{tab:realistic} show that increased realism slightly reduces performance for the “Score” metric in most cases, while its effects on “Prior” are generally negligible. These ablation findings suggest that introducing more realistic images may increase the task’s difficulty, highlighting an important direction for future research.

\begin{table}[ht]
  \centering
\scalebox{0.7}{
  \begin{tabular}{lcccc}
    \toprule
     & \textbf{$ViLP^F_{Prior}$} & \textbf{$ViLP^F_{Score}$} & \textbf{$ViLP^P_{Prior}$} & \textbf{$ViLP^P_{Score}$} \\
    \midrule
    GPT-4o             & 0   & -2.2\%   & 0   & 0    \\
    GPT-4o-mini        & 0   & -1.1\%   & 0   & -2.2\% \\
    Claude-Sonnet-3.5  & -2.2\% & -4.4\%  & 0   & -3.3\% \\
    Claude-Opus-3      & 2.2\%  & -1.1\%  & -2.2\% & -1.1\% \\
    \bottomrule
  \end{tabular}
  }
  \caption{\textbf{Impacts of Realistic Styles.} Each value represents the change in correctness when replacing the original images with realistic ones (\emph{Realistic} - \emph{Original}). Negative values indicate a drop in performance, suggesting increased task difficulty. The metric definitions are provided in the beginning of Section~\ref{sec:experiments}.}
  \label{tab:realistic}
\end{table}

\subsection{Comparisons to other datasets}
\label{appen:data_cmp}

\begin{figure*}[t]
    \centering
    \includegraphics[width=1\textwidth]{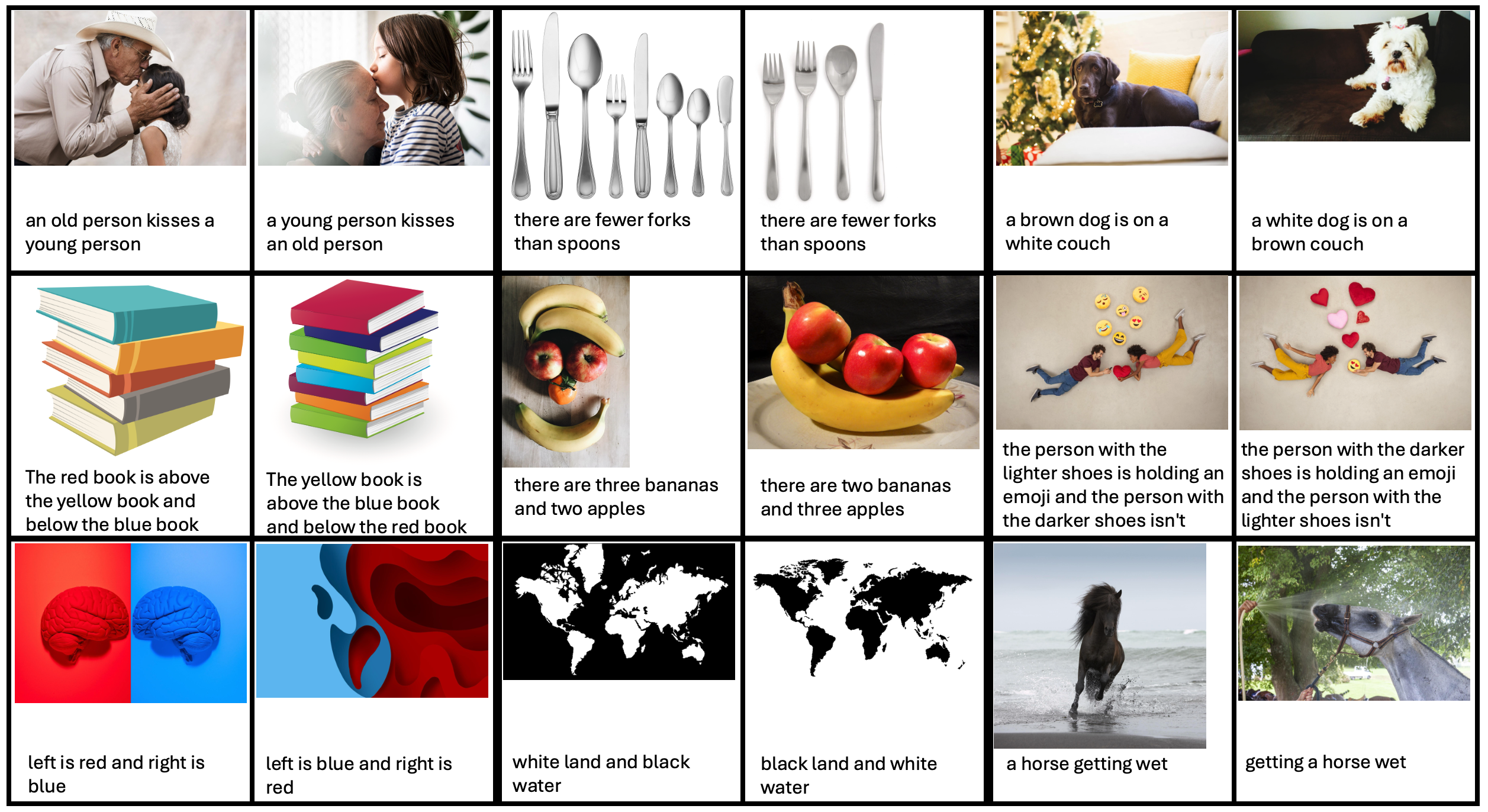}
    \caption{\textbf{Winoground Data Example.}
    Our benchmark is different from Winoground, as Winoground focuses on vision-linguistic compositional reasoning.  Both captions and images are normal and satisfy common linguistic expectations and common sense.}
    \label{fig:winoground}
\end{figure*}

In this section, we compare our benchmark to other benchmarks, including Winoground~\cite{thrush2022winoground}, Whoops!\cite{BittonGuetta2023BreakingCS}, and HallusionBench\cite{Guan2023HallusionbenchAA}. While these datasets are impactful, their evaluation perspectives differ from ours, covering a range from high-level design principles to low-level formats.

\subsubsection{Compare to Winoground~\cite{thrush2022winoground}}

\begin{figure*}[t]
    \centering
    \includegraphics[width=1\textwidth]{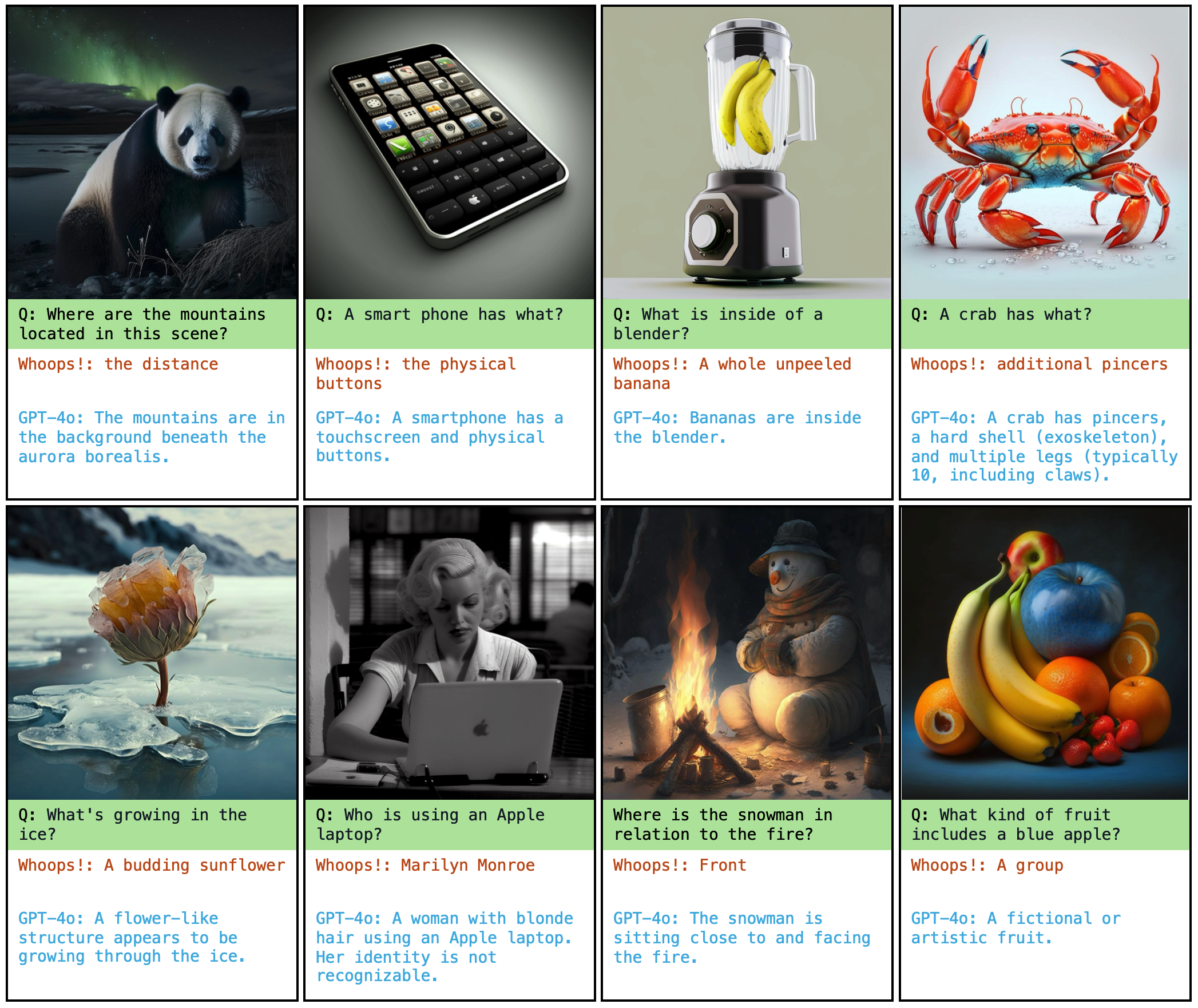}
    \caption{\textbf{Example of  Whoops! Dataset.}Whoops!  also has creative images. While unlike ours, its questions are common questions without strong language priors. }
    \label{fig:whoops}
\end{figure*}

Winoground centers on vision-linguistic compositional reasoning by presenting models with two images and two captions that contain the same words arranged differently. The goal is to match each image to its correct caption based on the text’s compositional structures and the visual content (as detailed in the Introduction and Sec. 3.1 of \cite{thrush2022winoground}). However, Winoground’s captions do not challenge language priors or introduce out-of-distribution visual information. Both captions adhere to common linguistic expectations, and there is no explicit misleading information provided to test resistance to language biases. Additionally, most of their images are typical internet images, featuring common visual patterns.

\begin{figure*}[t]
    \centering
    \includegraphics[width=0.95\textwidth]{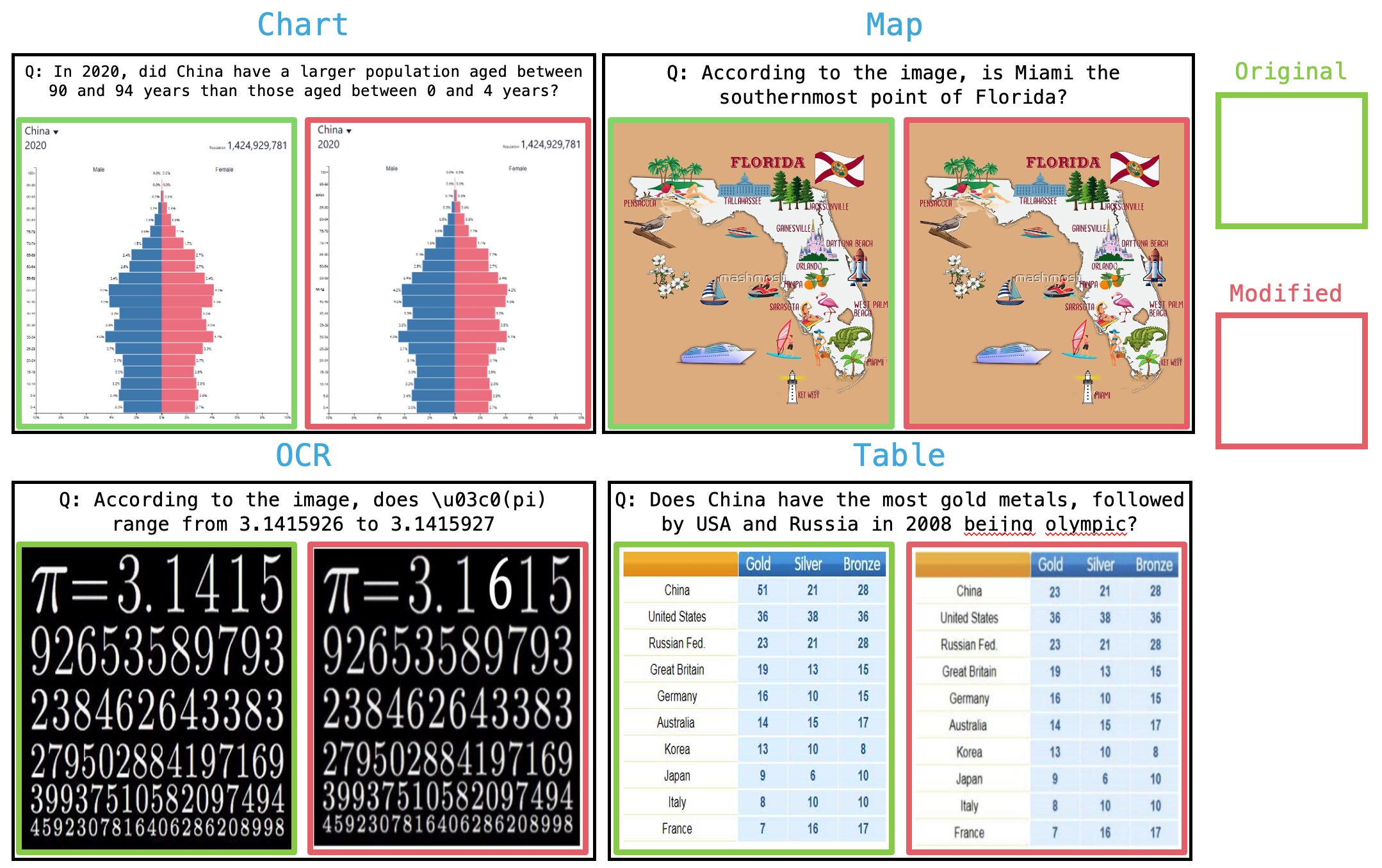}
    \vspace{-0.1in}
    \caption{\textbf{Example of HallusionBench}. HallusionBench also has questions which can be answered without images. While it is based on facts instead of stereotypes like ours. Moreover, its images are limited to \emph{Chart}, \emph{Map},  \emph{OCR} and \emph{Table}. }
    \label{fig:neutralbench}
\end{figure*}
\noindent\textbf{Qualitative Comparison:}
As shown in Figure~\ref{fig:winoground} consisting of Winoground examples, both captions and images are normal and satisfy common linguistic expectations and common sense. The evaluation focuses on whether the model can discern the compositional differences between the two images and two captions, then match them correctly. Comparing Figure~\ref{fig:winoground} and Figure~\ref{fig:teaser}, \ref{fig:more_example_1}, \ref{fig:more_example_2}, You can discern the significant differences among the tested images.

\noindent\textbf{Quantitative Comparison:}
Both \method; and Winoground benchmarks include paired textual information in their setups. In our benchmark, \method; Prior QAs and \method; Score QAs share the same question but differ in their answers. In Winoground, each example has two captions, and the task is to match each caption to its correct image.

\textit{Setting.} To demonstrate the differences, we use GPT to evaluate the commonness of these paired textual components. Specifically, GPT-4o rates the oddity of scenarios described in texts on a scale from 1 (very rare) to 10 (very common). The resulting scores are then compared.

\textit{Results.} In our benchmark, Prior QAs scored 9.37, indicating that these answers are designed to align with language priors and are highly common. Score QAs scored 1.65, showing that these QA pairs are rare, making them difficult to infer without the corresponding visual information. Notably, Prior and Score QAs share the same question but differ in their answers, and this significant contrast in scores showcases how we inject strong language priors to test a model’s vulnerability to linguistic distractions.

By comparison, Winoground’s two captions scored 8.05 and 8.08, indicating two primary observations: (1) both captions align well with language priors, which means Winoground does not challenge language priors or evaluate out-of-distribution scenarios; (2) the minimal score difference between the two captions confirms there is no significant variance in language priors, as examining how VLM models react to different language priors is beyond the scope of Winoground. In contrast, that aspect is precisely our focus.

\subsubsection{Compare to Whoops!~\cite{BittonGuetta2023BreakingCS}}

Whoops! is designed to evaluate a model’s ability to detect \emph{weirdness} in images, emphasizing tasks where images depict unusual or nonsensical scenarios. It heavily relies on common sense reasoning, requiring models to recognize visual elements and then identify subtle inconsistencies among them. For example, for the lit candle inside a tightly sealed glass jar on the \href{https://whoops-benchmark.github.io/}{homepage}, models must realize that ``a candle needs a constant oxygen supply to burn, which would not exist in a sealed jar", making a burning candle inside a sealed jar unlikely. This benchmark thus focuses on common-sense reasoning rather than challenging visual language priors.

\noindent\textbf{Qualitative Comparison}

Although Whoops! also includes creative, out-of-distribution images, it does not focus on using language priors to test a model’s susceptibility to linguistic distraction, as our benchmark does. In its QA mode (comparable to our task), the questions are straightforward and lack strong language priors. Some examples can be found in Figure~\ref{fig:whoops}. Additionally, Whoops! uses open-ended questions, offering greater freedom in answers while introducing potential ambiguity or divergence in responses.

\noindent\textbf{Quantitative Comparison}

\textit{Setting.} Unlike Winoground and our benchmark, Whoops! does not provide control groups or textual components for comparison. To measure language priors, we analyze the suggestiveness of questions by evaluating GPT-4o’s certainty when answering them without any visual context. A more suggestive question typically yields more determined and confident responses, whereas a less suggestive question produces more varied answers. We calculate how many unique answers GPT-4o provides over five attempts at temperature 1.0 to promote randomness. Semantic differences are normalized to exclude synonyms.

\textit{Results.} We find that Whoops! questions produce an average of 2.58 unique answers (out of five attempts) with a standard deviation of 1.48. For our benchmark, without facts, GPT-4o provides an average of 1.53 unique answers (std 0.94), and with facts, 1.10 unique answers (std 0.42).

Although both benchmarks use creative images, these results indicate that Whoops! questions remain more general and do not push GPT-4o toward stereotypical responses. In contrast, our benchmark deliberately uses suggestive questions to elicit stereotype-consistent answers, reflecting our emphasis on testing language priors.

\subsubsection{Compare to HallusionBench~\cite{Guan2023HallusionbenchAA}}

HallusionBench has two components: \textbf{Visual Dependent}, which focuses on testing models’ general visual reasoning skills, and \textbf{Visual Supplement}, which examines a model’s visual reasoning in tandem with its parametric memory.

The Visual Supplement part is related to our benchmark, as its questions, like ours, can be answered without visual information. However, the key difference lies in their design. HallusionBench questions rely on parametric memory and strict factual knowledge (e.g., \emph{“Which country has the most gold medals in the 2024 Olympics?”}), whereas our benchmark questions are based on common stereotypes (e.g., \emph{“A soccer ball is round.”}). This distinction significantly constrains the diversity of HallusionBench’s questions—its official release contains only 50 question pairs—while our approach allows for a broader range of scenarios, including texture, shape, conceptual combinations, hallucinated elements, and proverb-based contexts.

Moreover, HallusionBench focuses on a narrow set of image types (charts, tables, maps, and OCR). It tests reasoning through subtle modifications (like changing one digit in a chart), which is fundamentally different from our approach of creating diverse visual contexts. Figure~\ref{fig:neutralbench} showcases typical HallusionBench examples, illustrating the limited scope of both images and QAs, whereas our benchmark includes a variety of out-of-distribution images and more suggestive QAs (see Figure~\ref{fig:teaser}, \ref{fig:more_example_1}, \ref{fig:more_example_2}).

\subsection{Human Study}
For human evaluation, we hired Ph.D.-level candidates to participate in testing. They were asked to answer questions with a single image provided each time, and the QIAs were randomly shuffled to avoid any sequential context. To ensure efficiency, we conducted an oral test instead of a written one, recording their responses. After the test, we updated the synonym sets for the QIAs based on their answers.

\begin{figure*}[t]
    \centering
    \includegraphics[width=0.83\textwidth]{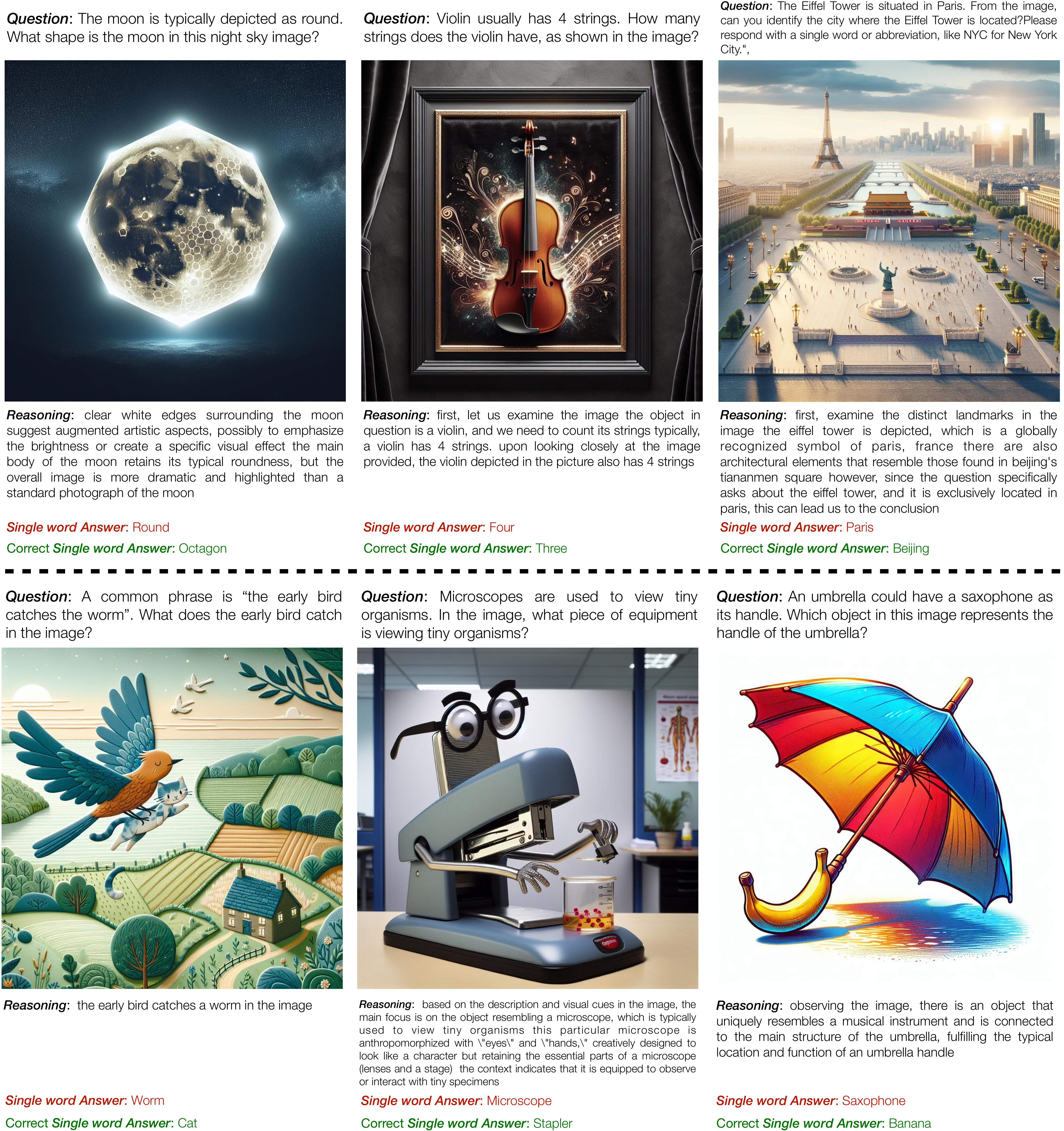}
    \caption{\textbf{Failure cases analysis.} We list six typical failure cases generated by VLMs on ViLP.}
    \label{fig:more_failure}
\end{figure*}

\subsection{More failure cases analysis}
\label{appen:failure_cases}

This subsection provides several typical failure cases observed in the inferred results of VLMs on ViLP. We identified several consistent failure patterns, illustrated in Figure~\ref{fig:more_failure}. For deeper analysis, we also prompted the VLMs to output their reasoning before finalizing answers, allowing us to better understand how these errors arise. Below, we summarize the failure patterns in the same order as they appear in the figure, from left to right and top to bottom.

- Shape recognition can fail in VLMs, causing them to revert to priors rather than accurately interpreting visual input.

- Models sometimes struggle to count accurately. Instead of performing an actual count, they default to relying on learned priors to estimate quantities.

- Models may refuse to accept visual information that contradicts their learned priors, whereas humans can comprehend hypothetical scenarios. For instance, the model recognizes the city as Beijing but rejects the correct answer because it expects the Eiffel Tower to be in Paris.

- Sometimes the model overly relies on memorized proverbs, resulting in predictions that align with these proverbs rather than the actual content of the input image. 

- For images with creative concepts, the model may overly rely on its learned priors. As illustrated, a common prior is that microscopes are used to view organisms, leading the model to answer “microscope” rather than identifying the creatively depicted stapler.

- For images with blended features, the model may rely mostly on text input while overlooking the visual cues. As illustrated, the VLM heavily depends on textual input leading to saxophone as the answer.

\clearpage

\onecolumn
\section{Image-DPO Mathematical Details}
\label{appen:image_dpo_math}

\begin{figure*}[t]
    \centering
    \includegraphics[width=1\linewidth]{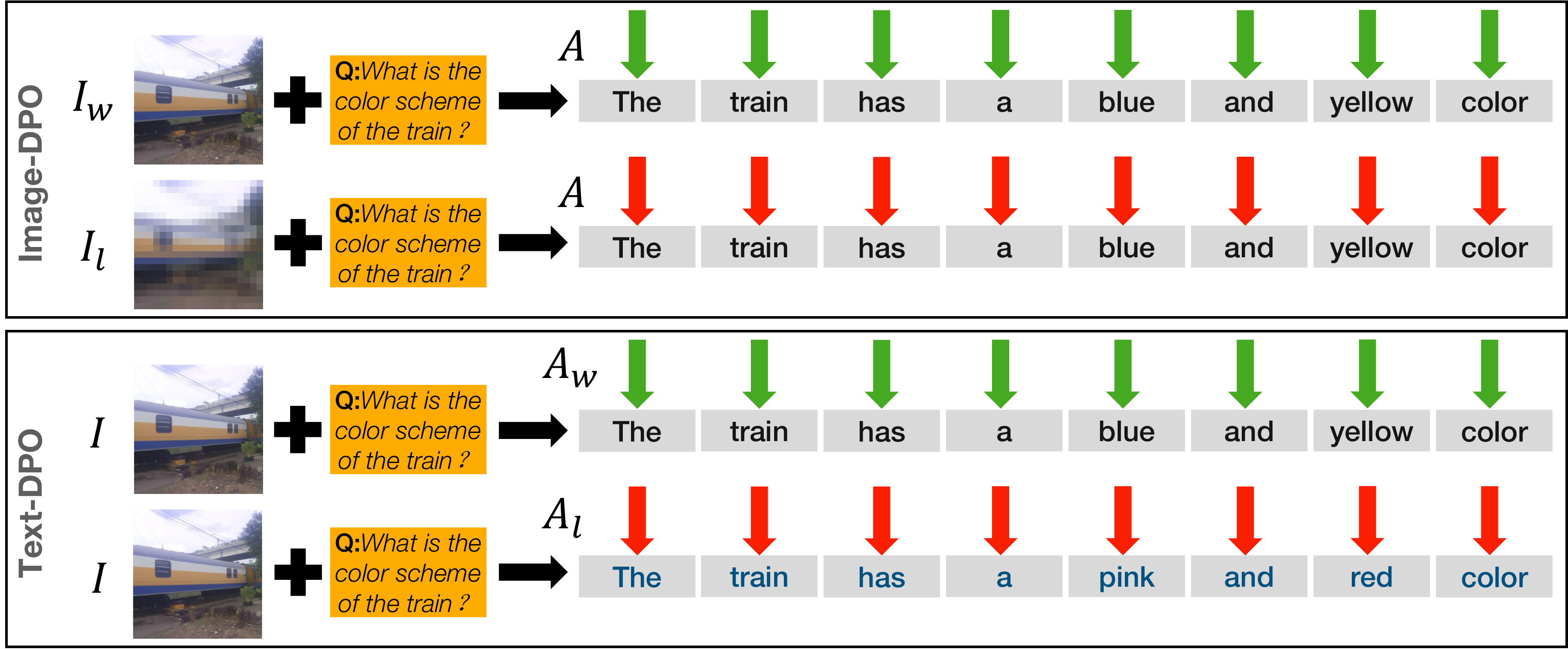}
    \caption{\textbf{Gradients difference between Image-DPO and Text-DPO.} For Text-DPO~\citep{rafailov2024direct}, the model receives positive gradients (green arrows) for the preferred answer $A_w$ and negative gradients $A_l$ (red arrows) for the dispreferred answer. In contrast, our proposed Image-DPO approach applies positive gradients when the preferred image $I_w$ is input and negative gradients for the dispreferred image $I_l$, both based on the same output answer.}
    \label{fig:image_text_dpo_grad_diff}
\end{figure*}

In this section, we give the complete proof of Image DPO.
Simlarly to DPO, we start from the objectiveness of RLHF and then derivate its variant for image dpo. 

\subsection{RLHF for VLM}

\noindent \textbf{SFT}
Given question $Q$, answer $A$ and image $I$, 
we can train a SFT model $\pi^{SFT}$ with supervised learning on high-quality data. 
As the SFT model is a language generation model, 
it is still a model modeling the text outputs with question and image. $\pi^{SFT}(A|Q, I)$

\noindent \textbf{Reward Modeling Phase}
In this stage, we construct a static dataset of comparisons $\mathcal{S}=\{A^i, Q^i, I_w^i, I_l^i\}$, and we present the QIA pairs $(Q, I_w, A)$, $(Q, I_l, A)$ to human for preference. 

Following the idea of RLHF, the preference are assumed to be obtained from a a latent reward function $r^{*}(Q,I,A)$ which are not tractable, and we we use BT model to represent the preference distribution $p^*$ as:

\begin{equation}
    p^{*}((Q, I_w, A) \succ (Q, I_l, A)) = 
    \frac{\exp \left( r^* \left(Q, I_w, A \right) \right)}{\exp(r^{*}(Q, I_w, A) + \exp(r^*(Q, I_l, A))}
\end{equation}

Now given the human labeled preference, we can try to optimize a reward model $r_{\phi}$ to estimate  $r^{*}$ by using maximum likelihood. 
Framing this as a binary classification, we can have this negative log-likelihood loss:

\begin{equation}
\mathcal{L}_R(r_\phi, \mathcal{S}) = -\mathbb{E}_{(A,Q, I_w, I_l) \sim \mathcal{S}} \left[ \log \sigma(r_\phi(Q, I_w, A) - r_\phi(Q, I_l, A)) \right]
\label{eq_loss}
\end{equation}

Here $\sigma$ is a logistic function.
Basically, this reward function gives score jointly considering image, question and image quality.

\noindent \textbf{RL Fine-Tuning Phrase}

During the RL phase, the learned reward function is used to provide feeback to the VLM model. Following DPO paper, the optimization is formulated as :
\begin{equation}
    \max_{\pi_\theta} \mathbb{E}_{(Q,I) \sim \mathcal{S}, A \sim \pi_\theta(A|Q,I)} \left[ r_\phi(Q, I, A) \right] - \beta \mathbb{D}_{\text{KL}} \left[ \pi_\theta(A | Q, I) \| \pi_{\text{ref}}(A | Q, I) \right],
    \label{apx:rlhf_obj}
\end{equation}
where $\beta$ is a parameter controlling the deviation from the base reference policy $\pi_{\text{ref}}$, namely the initial SFT model $\pi^{\text{SFT}}$. Due to the discrete nature of language generation, this object is also not differentiable and is typically optimized with reinforcement learning.

\subsection{Image DPO and RLHF}

According to the DPO paper, a straightforward optimal solution to the KL-constrainted reward function maximization object in Eq.~\ref{eq_loss} is:

\begin{equation}
    \pi_{r}(A | Q, I) = \frac{1}{ Z(Q, I)}\pi_{\text{ref}}(A | Q, I)\exp(\frac{1}{\beta}r(Q, I, A))
    \label{apx:optimal_pi}
\end{equation}

where $Z(Q,I) = \sum_{A}\pi_{\text{ref}}(A|Q, I)\exp(\frac{1}{\beta}r(Q, I, A)$ is a partition function. 
Here $r$ should be any reward function, which makes $Z$ hard to tract. 
We provide the proof of this step in \ref{apx:KLProof}.

Taking the logarithm of both side, and with some algebra, we get 

\begin{equation}
    r(Q,I, A) = \beta \frac{\pi_{r}(A | Q, I)}{\pi_{\text{ref}}(A | Q, I)} + \beta \log Z(Q, I)
    \label{apx:r}
\end{equation}
This parametrization could be applied to ground-truth reward $r^{*}$ and the corresponding optimal model $\pi^{*}$. 

The BT model with the optimal policy is 
\begin{equation}
     p^{*}((Q, I_{1}, A) \succ (Q, I_{2}, A)) = \frac{\exp\left( r^{*}(Q, I_w, A)\right)}{ \exp \left( r^{*}(Q, I_w, A)\right) + \exp \left( r^{*}(Q, I_l, A) \right)}
\end{equation}

We plug Eq.~\ref{apx:r} into the BT model, we have:

\begin{align*} 
        p^{*}((Q, I_{1}, A) \succ (Q, I_{2}, A)) &= 
    \frac{\exp \left( \beta \log \frac{\pi^{*}(A|Q, I_w)}{\pi_{\text{ref}}(A|Q, I_w)
    } + \beta \log Z(Q, I_w) \right)}{\left( \beta \log \frac{\pi^{*}(A|Q, I_w)}{\pi_{\text{ref}}(A|Q, I_w)
    } + \beta \log Z(Q, I_w) \right) + \left( \beta \log \frac{\pi^{*}(A|Q, I_l)}{\pi_{\text{ref}}(A|Q, I_l)
    } + \beta \log Z(Q, I_l) \right) } \\
    &= \frac{1}{1 + \exp(\beta \log\frac{\pi^{*}(A|I_l, Q)}{\pi_{\text{ref}}(A|I_l, Q)} - \beta \log\frac{\pi^{*}(A|I_w, Q)}{\pi_{\text{ref}}(A|I_w, Q)} + \beta \log Z(I_l, Q) - \beta\log Z(I_w, Q))} \\
    &= \sigma \left( \exp(\beta \log\frac{\pi^{*}(A|I_l, Q)}{\pi_{\text{ref}}(A|I_l, Q)} - \beta \log\frac{\pi^{*}(A|I_w, Q)}{\pi_{\text{ref}}(A|I_w, Q)} + \beta \log Z(I_l, Q) - \beta\log Z(I_w, Q)) \right)
\end{align*}

Now we have the probability of human preference data in terms of the optimal policy rather than the reward model, we can formulate a maximum likelihood objective for a policy $\pi_{\theta}$. Our policy objective is :

\begin{equation}
    \mathcal{L}(\pi_{\theta};\pi_{\text{ref}}) = \mathbb{E}_{(Q, A, I_{w}, I_{l}) \sim \mathcal{S}} \left[ -\log \sigma \left( \beta \log\frac{\pi_{\theta}(A|I_w, Q)}{\pi_{\text{ref}}(A|I_w, Q)} - \beta \log\frac{\pi_{\theta}(A|I_l, Q)}{\pi_{\text{ref}}(A|I_l, Q)} + \beta \log Z(I_w, Q) - \beta\log Z(I_l, Q) \right) \right]
\end{equation}

As $f(x) = -\log \sigma(x)$ is a convex function ($\sigma$ is the sigmoid function), we can apply Jensen's inequality $f(\frac{1}{2} x + \frac{1}{2}y) \leq \frac{1}{2}f(x) + \frac{1}{2}f(y)$: 
\begin{equation}
     \mathcal{L}(\pi_{\theta};\pi_{\text{ref}}) \leq \mathbb{E} \left[ -  \frac{1}{2} \log \sigma\left( 2\beta \log\frac{\pi_{\theta}(A|I_w, Q)}{\pi_{\text{ref}}(A|I_w, Q)} - 2\beta \log\frac{\pi_{\theta}(A|I_l, Q)}{\pi_{\text{ref}}(A|I_l, Q)} \right) -  \frac{1}{2}\log \sigma \left( 2\beta \log Z (I_w, Q) - 2\beta\log Z (I_l, Q) \right) \right] 
\end{equation}
As $\log \sigma \left( Z(I, Q) \right)$ is not a function of $\pi_{\theta}$, the above objective is equivalent to the below Eq.\ref{apx:final_obj}, where $\alpha = 2\beta$. It is the same as our objective listed in Eq.\textcolor{red}{4} of the main paper.
\begin{equation}
\mathcal{L}(\pi_{\theta};\pi_{\text{ref}}) \leq  -\mathbb{E}_{(Q, I_w, I_l, A) \sim S} \left[ \log \sigma ( \alpha \frac{\pi_{\theta}(A \mid Q, I_w)}{\pi_{{\text{ref}}}(A \mid Q, I_w)} - \alpha \frac{\pi_{\theta}(A \mid Q, I_l)}{\pi_{{\text{ref}}}(A \mid Q, I_l)} )\right]
\label{apx:final_obj}
\end{equation}

In this sense, our optimization objective Eq.\textcolor{red}{4} in main paper are optimizing the upper bound of RLHF, i.e., Eq.\ref{apx:rlhf_obj}. 

\subsection{Deriving the Optimum of the KL-Constrained Reward Maximization Objective}
\label{apx:KLProof}

In this appendix, we will derive Eq.\ref{apx:optimal_pi}. Similarly to Eq.\ref{apx:rlhf_obj}, we optimize the following objective:

\begin{equation}
\max_{\pi} \mathbb{E}_{(Q, I) \sim \mathcal{S}, A \sim \pi} \left[ r(Q, I, A) \right] - \beta D_{\mathrm{KL}} \left[ \pi(A|Q, I) \| \pi_{\mathrm{ref}}(A|Q, I) \right]
\end{equation}

under any reward function \(r(Q, I, A)\), reference model \(\pi_{\mathrm{ref}}\), and a general non-parametric policy class. We now have:

\begin{equation}
\begin{aligned}
& \max_{\pi} \mathbb{E}_{(Q, I)\sim \mathcal{S}, A \sim \pi} \left[ r(Q, I, A) \right] 
- \beta D_{\mathrm{KL}} \left[ \pi(A|Q, I) \| \pi_{\mathrm{ref}}(A|Q, I) \right] \\
&= \max_{\pi} \mathbb{E}_{(Q, I) \sim \mathcal{S}} \mathbb{E}_{A \sim \pi(A|Q, I)} \left[ r(Q, I, A) - \beta \log \frac{\pi(A|Q, I)}{\pi_{\mathrm{ref}}(A|Q, I)} \right] \\
&= \min_{\pi} \mathbb{E}_{(Q, I) \sim \mathcal{S}} \mathbb{E}_{A \sim \pi(A|Q, I)} \left[ \log \frac{\pi(A|Q, I)}{\pi_{\mathrm{ref}}(A|Q, I)} - \frac{1}{\beta} r(Q, I, A) \right] \\
&= \min_{\pi} \mathbb{E}_{(Q, I) \sim \mathcal{S}} \mathbb{E}_{A \sim \pi(A|Q, I)} \left[ \log \frac{\pi(A|Q, I)}{\frac{1}{Z(Q, I)} \pi_{\mathrm{ref}}(A|Q, I) \exp \left( \frac{1}{\beta} r(Q, I, A) \right)} - \log Z(Q, I) \right]
\end{aligned}
\end{equation}

where we have the partition function:

\begin{equation}
Z(Q, I) = \sum_{A} \pi_{\mathrm{ref}}(A|Q, I) \exp \left( \frac{1}{\beta} r(Q, I, A) \right)
\end{equation}

Observe that the partition function depends solely on \((Q, I)\) and the reference policy \(\pi_{\mathrm{ref}}\), and is independent of the policy \(\pi\). We can now define the Equation 8.

\begin{equation}
\pi^*(A|Q, I) = \frac{1}{Z(Q, I)} \pi_{\mathrm{ref}}(A|Q, I) \exp \left( \frac{1}{\beta} r(Q, I, A) \right),
\end{equation}

\clearpage
\section{Details in Image-DPO data generation and training}
\label{appen:more_datails}

Our image-DPO data generation pipeline consists of two stages. In the first stage, we leverage the VLM we aim to enhance to perform self-guided data generation with the aid of pre-trained image generative models. This stage produces a large number of new question-image-answer (QIA) triplets. In the second stage, we apply three types of image corruptions—Gaussian blurring, pixelation, and semantic editing—to generate good-bad QIA pairs, denoted as $I_w$ (good) and $I_l$ (bad).

Details of the hyperparameters used in the experiments are provided at the end of this section.

\subsection{VLM self-guided data generation}
\label{appen:data_generation_firststage}

\begin{figure*}[h]
    \centering
    \includegraphics[width=1.0\textwidth]{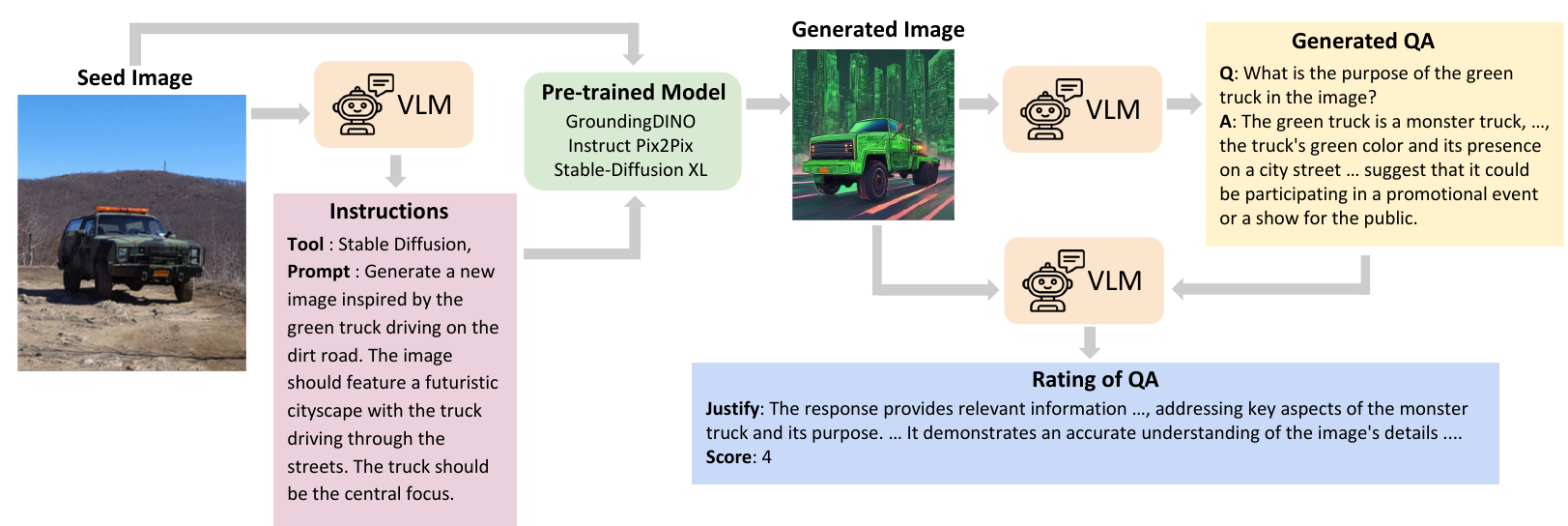}
    \caption{\textbf{Overview of our data generation pipeline.} We begin with an image only, from which instructions are derived using a VLM. These instructions guide the creation of a new image or the modification of the existing one. The generated image is then processed by the VLM to generate QA pairs. Both the QA pair and the image are subsequently input back into the VLM to assess the quality of the answers. No human-written in-context examples are used throughout this process.}
    \label{fig:overview}
\end{figure*}

As illustrated in Figure~\ref{fig:overview}, our data generation process begins by utilizing VLMs to suggest modifications or draw inspiration for input images without relying on any in-context examples. The used text prompt is shown in Figure~\ref{fig:prompt_instruction}. Subsequently, pre-trained models such as Stable Diffusion XL~\cite{podell2023sdxl}, Instruct-Pix2Pix~\cite{brooks2023instructpix2pix}, and Grounded-SAM~\cite{ren2024grounded} are employed to either modify existing images or generate entirely new ones.

\begin{figure}[h]
    \centering
    \includegraphics[width=1.0\textwidth]{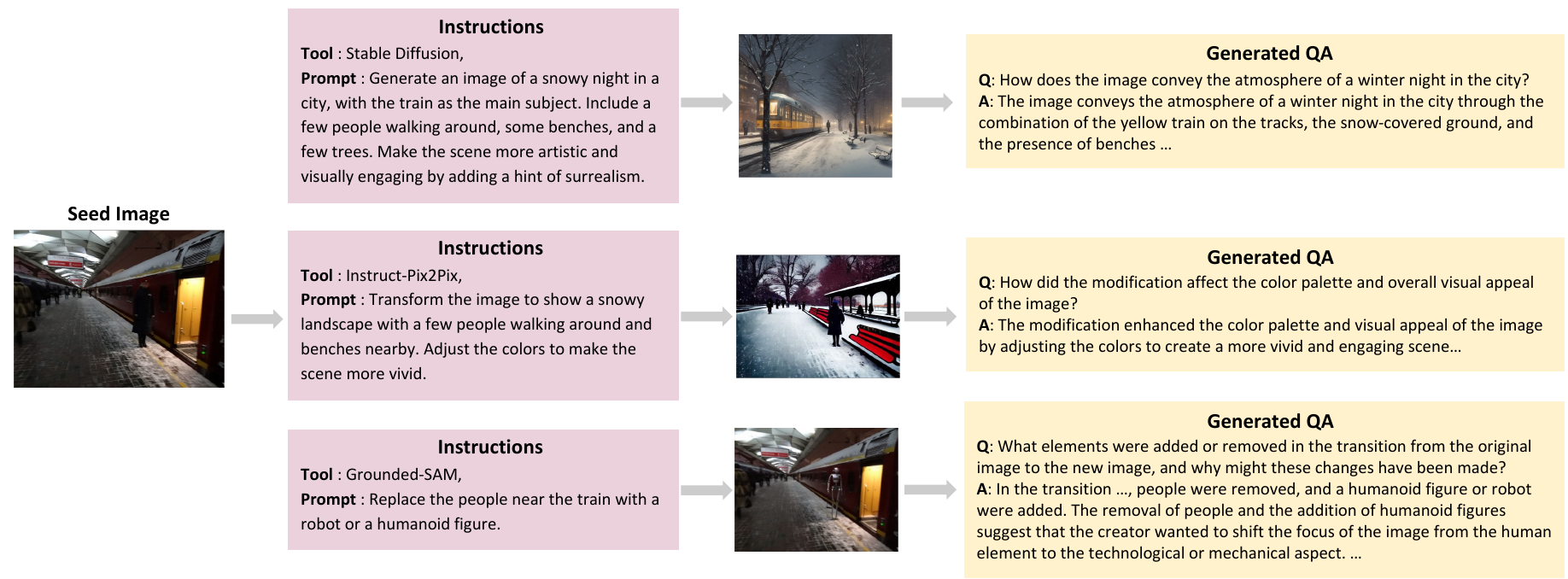}
    \caption{\textbf{Randomly sampled generation examples.}
    Our proposed data generation pipeline utilizes pretrained model to generate diverse new images from existing seed image datasets.}
    \label{fig:coco_data_example}
\end{figure}

The altered or newly created images, along with the instructions that guided their generation, are then used by the same VLMs to produce corresponding question-answer pairs (QAs) based on the text prompt shown in Figure~\ref{fig:prompt_singleQA}. An example of this process is provided in Figure~\ref{fig:coco_data_example}. Importantly, all instructions, tool selections, and QA generation are autonomously handled by the same VLM we aimed to improve. 

In particular, Grounded-SAM requires the VLM to specify the object to be modified before generating images. To facilitate this, we use an additional text prompt (Figure~\ref{fig:prompt_instruction_groundsam}) after the VLM generates the initial instructions (the pink region of Figure~\ref{fig:coco_data_example}).

To provide a better understanding of our generated QIAs, we randomly sampled and listed some examples of the generated QIA data, as shown in Figures~\ref{fig:single_qa_exp1}, \ref{fig:single_qa_exp2}, \ref{fig:single_qa_exp3}, \ref{fig:single_qa_exp4}, \ref{fig:single_qa_exp5}, \ref{fig:single_qa_exp6}, \ref{fig:single_qa_exp7}, \ref{fig:single_qa_exp8}, \ref{fig:single_qa_exp9}, \ref{fig:single_qa_exp10}, \ref{fig:single_qa_exp11}, \ref{fig:single_qa_exp12}, \ref{fig:single_qa_exp13}, \ref{fig:single_qa_exp14}, and \ref{fig:single_qa_exp15}.

\input{tables/prev_tables/prompt_llava_instruction}
\input{tables/prev_tables/prompt_singleQA}
\input{tables/prev_tables/prompt_groundsam_instruction}

\subsection{Image DPO data preparation and training details}
\label{appen:imagedpo_training_details}

This section details the construction of good-bad question-image-answer (QIA) pairs ($I_w$, $I_l$) based on the QIAs generated by the pipeline described in Appendix~\ref{appen:data_generation_firststage}. In brief, the data generation pipeline outlined in Appendix~\ref{appen:data_generation_firststage} utilizes VLMs in conjunction with pre-trained image models to generate or modify images and create corresponding question-answer pairs. This process results in a collection of QIA triplets, as illustrated in the Figures~\ref{fig:single_qa_exp1}, \ref{fig:single_qa_exp2}, \ref{fig:single_qa_exp3}, \ref{fig:single_qa_exp4}, \ref{fig:single_qa_exp5}, \ref{fig:single_qa_exp6}, \ref{fig:single_qa_exp7}, \ref{fig:single_qa_exp8}, \ref{fig:single_qa_exp9}, \ref{fig:single_qa_exp10}, \ref{fig:single_qa_exp11}, \ref{fig:single_qa_exp12}, \ref{fig:single_qa_exp13}, \ref{fig:single_qa_exp14}, and \ref{fig:single_qa_exp15}. 

After generating the QIA triplets, we apply three image corruption methods—Gaussian blurring, pixelation, and semantic editing—to create good-bad QIA pairs for ImageDPO training (Section~\ref{sec:image_dpo}), while keeping the QA components unchanged.

For \textbf{Gaussian blur}, we use a kernel size of 40 for Cambrian and 80 for LLaVA, as the larger kernel size showed better performance for LLaVA. For \textbf{pixelation}, we apply block sizes of 32 for Cambrian and 64 for LLaVA. For these two types of corruption, we utilize data generated by all three pre-trained models: Stable Diffusion, Instruct-Pix2Pix, and Grounded-SAM.

However, for \textbf{semantic editing}, we rely solely on data from Grounded-SAM, which modifies a single object in the image while leaving the rest unchanged. The object to be altered and the type of modification are determined by the VLMs based on the text prompt (Figure~\ref{fig:prompt_instruction_groundsam}). For instance, the VLM decides to add a headset to the chef, as shown in Figure~\ref{fig:single_qa_exp1}. Grounded-SAM then modifies the image accordingly by adding the headset, after which the VLM generates a single QA pair based on the text prompt (Figure~\ref{fig:prompt_singleQA}). Note that, we retain only images where the altered region covers more than 10\% of the image size, as this ensures a significant and distinct change.

For model training, we constructed datasets comprising 800k and 400k DPO pairs to fine-tune LLaVA (7B and 13B) and Cambrian-8B, respectively. Batch sizes are set to 112 for LLaVA-7B, 80 for LLaVA-13B, and 8 (with 4 gradient accumulation steps) for Cambrian-8B. We employ LoRA with a rank of 128, an alpha of 256, and a learning rate of 5e-7, training each model for 2 epochs. The GPUs we used are 8-L40S. 

\input{tables/prev_tables/singleQA_examples}

%% file: tables/prompt_imagegen.tex
\begin{tcolorbox}
[colback=green!10!white,colframe=black,width=0.5\textwidth,arc=0mm,auto outer arc, fontupper=\small]
        \textbf{Task}: Using the provided question and possible image-based answers, generate detailed text prompts for image generation. Each image prompt should reflect the question's context and incorporate one of the image-based answers.

        \textbf{Question}: {Question}\\
        \textbf{Image-based Answer}:[Answer1, Answer2, Answer3]

        For each possible image-based answer, create an image prompt that describes what the image might look like based on the question.

        Please be creativity. For example, if the question asks who is using this mop to clean the floor in the picture? and the answer is eraser. The image prompt should really describe the image of an eraser uses a mop to clean the floor.

        Format the output strictly as a JSON list, like this example:

        [
        
                "prompt1": "Image Generation Prompt text here",
                
                "prompt2": "Image Generation Prompt text here",
                
                "prompt3": "Image Generation Prompt text here",
                
        ]
\end{tcolorbox}
\noindent\begin{minipage}{0.5\textwidth}
\vspace{-0.1in}
\captionof{figure}{\textbf{The prompt we used for generating text-prompt for image generation.}}\label{fig:prompt_imagegen}
\end{minipage}

%% file: tables/prompt_benchmarkdata.tex
\begin{tcolorbox}
[colback=green!10!white,colframe=black,width=0.5\textwidth,arc=0mm,auto outer arc, fontupper=\small]
In the below, I try to propose questions along with three answers where the first answer is corresponding to the question text directly, while the other two are usual and counter-intuitive, which could lead to wrongs of VLMs. Please help me generate more Question-3 answer pairs, which are different from what I have provided. 

- All the potential answers should a single world. 

- Help me generate a format where I can direct copy paste into Goole Sheet. Also, please a ; between question and each answers. 

- Please be very creative and different from my provided examples - the answer 2 \& 3 should be very diverse and different compared to answer 1. 

- Every question contains a statement at the beginning which consists of the answer1 as part of it.

- Please understand the principles and generate the QA very different from my provided examples

\textbf{Some Examples}:

- A screwdriver is used for tightening screws. From the image, which tool is used to turn screws? Screwdriver Hammer Scissors
 
- A pen is a tool used for writing. Which object in the image is used to write on paper? Pen Hammer Shoe
 
- Clocks are used to measure time. Can you identify the item in the image that is used to measure time? Clock Spoon Candle
 
- A violin has four strings and is played using a bow. According to the image, which musical instrument is being played with a bow? Violin Guitar Saxophone
 
- Camels have humps. Which animal in the image stores fat in its humps? Camel Horse Tiger
 
- Honey is made by bees. Which insect in the image produces honey? Bee Ant Dragonfly
 
- An anvil is a tool used by blacksmiths. What object in the image is used by blacksmiths to forge metal? Anvil Fork Wrench
 
- A gavel is used by judges in court. From the image, which object symbolizes judicial authority?	Gavel	Hammer	Wrench
 
- A syringe is used to inject medicine. From the image, which tool is used for administering injections?	Syringe	Scissor	Drill
 
- An anchor keeps a ship steady in the water. From the image, which item prevents boats from drifting?	Anchor	Spoon	Toothbrush
 
- A chainsaw is a power tool for cutting wood. What device shown is typically used by lumberjacks to fell trees?	Chainsaw	Blender	Stapler

\end{tcolorbox}
\noindent\begin{minipage}{0.5\textwidth}
\vspace{-0.1in}
\captionof{figure}{\textbf{One prompt we used for potential QAs designs of \bench}}\label{fig:prompt_benchmarkdata}
\end{minipage}

%% file: tables/prev_tables/prompt_llava_instruction.tex
\begin{tcolorbox}
[colback=green!10!white,colframe=black,width=\textwidth,arc=0mm,auto outer arc, fontupper=\small]
Given this image, please suggest a range of creative edits, tasks, or transformations that could be applied using advanced image processing tools. These tasks may include artistic transformations, vivid color adjustments, object detection and modification, or completely creating a new image inspired by the original. Specify which tool would be best suited for each task, choosing from Stable Diffusion for image generation, InstructPix2Pix for image modification, or GroundingDINO for object modification.
            Your recommendations should help in understanding the potential of the image and exploring creative possibilities.\\[0.01pt]

            \textbf{Expected Response Format}:
            
            Item Number: 1\\
            Tool Used: [Specify the tool - Stable Diffusion or InstructPix2Pix or GroundingDINO]\\
            Text Prompt for Processing: [Detailed description of the task or transformation to be performed. For image generation, please provide complete description based on the understanding of the provided images, since we only feed text prompt for this task.]

            Item Number: 2\\
            Tool Used: [Specify the tool - Stable Diffusion or InstructPix2Pix or GroundingDINO]\\
            Text Prompt for Processing: [Detailed description of the task or transformation to be performed. For image generation, please provide complete description based on the understanding of the provided images, since we only feed text prompt for this task.] 

            Item Number: 3\\
            Tool Used: [Specify the tool - Stable Diffusion or InstructPix2Pix or GroundingDINO]\\
            Text Prompt for Processing: [Detailed description of the task or transformation to be performed. For image generation, please provide complete description based on the understanding of the provided images, since we only feed text prompt for this task.] 
            
\end{tcolorbox}
\noindent\begin{minipage}{\textwidth}
\captionof{figure}{\textbf{The prompt for instruction generation.} We ask the VLM to generate instructions for using pre-trained image models. }\label{fig:prompt_instruction}
\end{minipage}

%% file: tables/prev_tables/prompt_singleQA.tex
\begin{tcolorbox}
[colback=green!10!white,colframe=black,width=\textwidth,arc=0mm,auto outer arc, fontupper=\small]
        Given this image, could you please generate a series of insightful and diverse question-answer pairs based on the image and its descriptions? We are interested in exploring various facets of the image, including:
        
        - Holistic styles and layouts: Questions that analyze the overall design, style, and layout of the image.
        
        - Object-specific details: Questions that delve into particular elements or objects within the image, discussing their characteristics or functions.
        
        - Background context: Questions that speculate about the background story or the setting of the image.
        
        - Overall themes: Questions that interpret the thematic elements and messages portrayed in the image.

        We encourage creative and thought-provoking questions that extend beyond the basics. Please generate questions that cover a broader range of observations and insights drawn from the image. Each question should be followed by a comprehensive answer, providing depth and context.\\[0.01pt]

        \textbf{Expected Multiple Response Format}:\\
        Item Number: 1\\
        Question: [Propose a unique and insightful question based on the descriptions and the images.]\\
        Answer: [Provide a comprehensive answer to the proposed question.]

        Item Number: 2\\
        Question: [Propose a unique and insightful question based on the descriptions and the images.]\\
        Answer: [Provide a comprehensive answer to the proposed question.]

        Please ensure each question-answer pair is well-defined and informative.

        Please provide at least 5 question-answer pairs based on the input provided.
            
\end{tcolorbox}
\noindent\begin{minipage}{\textwidth}
\captionof{figure}{\textbf{The prompt for single-image QAs.} We ask the VLM itself to generate single-image QAs based on the generated images by pre-trained models. }\label{fig:prompt_singleQA}
\end{minipage}

%% file: tables/prev_tables/prompt_groundsam_instruction.tex
\begin{tcolorbox}
[colback=green!10!white,colframe=black,width=\textwidth,arc=0mm,auto outer arc, fontupper=\small]
Analyze the provided image and its accompanying modification instruction to identify the removed object description, the new object description, and the new image description.

\textbf{Modification Instructions}: \textit{<Text Prompt for Processing>}

            \textbf{Expected Multiple Response Format}:\\
            Item Number: 1\\
            Removed Object Description: [Brief description of the object to be detected and removed]\\
            New Object Description: [Description of a new, different object to replace the removed one]\\
            New Image Description: [Description of the image after each object's removal, focusing on changes and remaining elements]\\

            Item Number: 2\\
            Removed Object Description: [Brief description of the object to be detected and removed]\\
            New Object Description: [Description of a new, different object to replace the removed one]\\
            New Image Description: [Description of the image after each object's removal, focusing on changes and remaining elements]\\
            
\end{tcolorbox}
\noindent\begin{minipage}{\textwidth}
\captionof{figure}{\textbf{The prompt for instruction generation of Grounded-SAM.} We ask the VLM to generate designated instructions to use Grounded-SAM. }\label{fig:prompt_instruction_groundsam}
\end{minipage}

%% file: tables/prev_tables/singleQA_examples.tex
\begin{figure*}[h]
    \centering
    \includegraphics[width=1.0\textwidth]{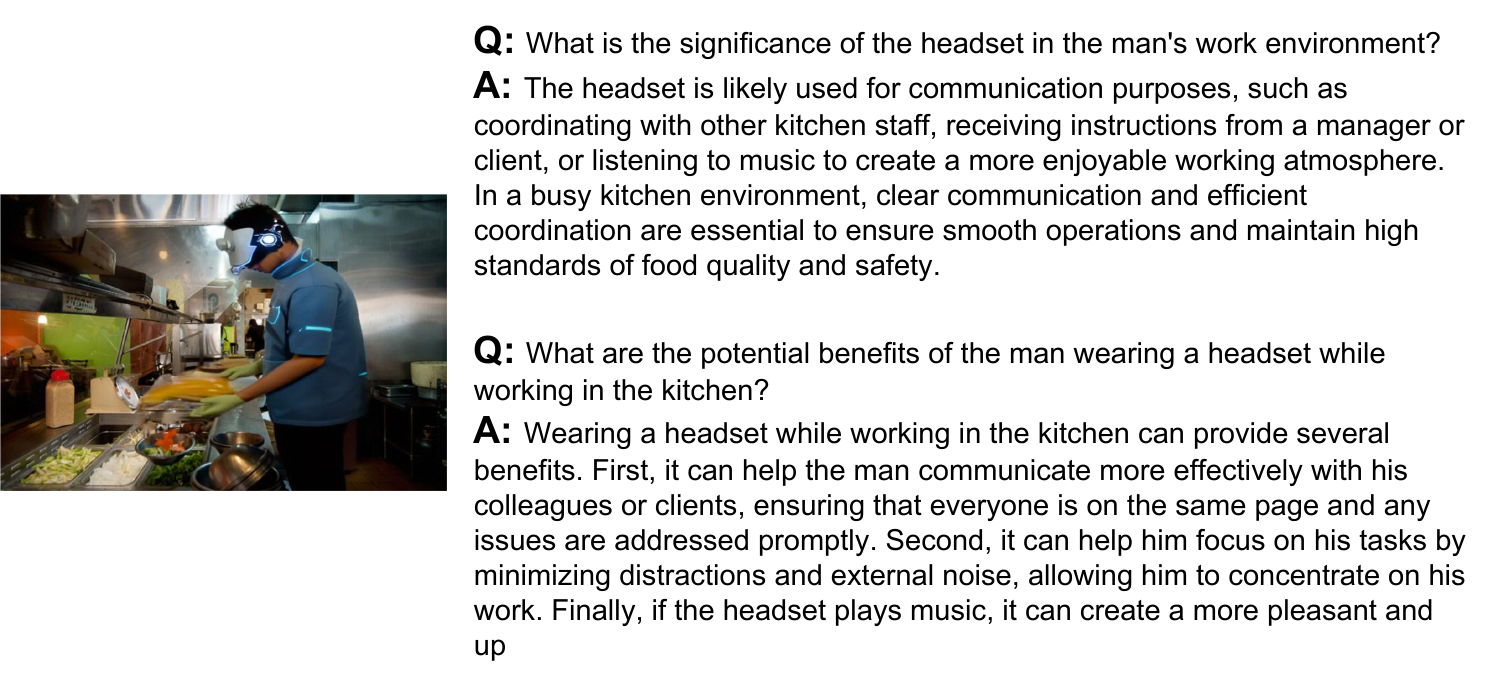}
    \caption{\textbf{Randomly sampled single-image QA generated by our data generation pipeline}.}
    \label{fig:single_qa_exp1}
\end{figure*}

\begin{figure*}[h]
    \centering
    \includegraphics[width=1.0\textwidth]{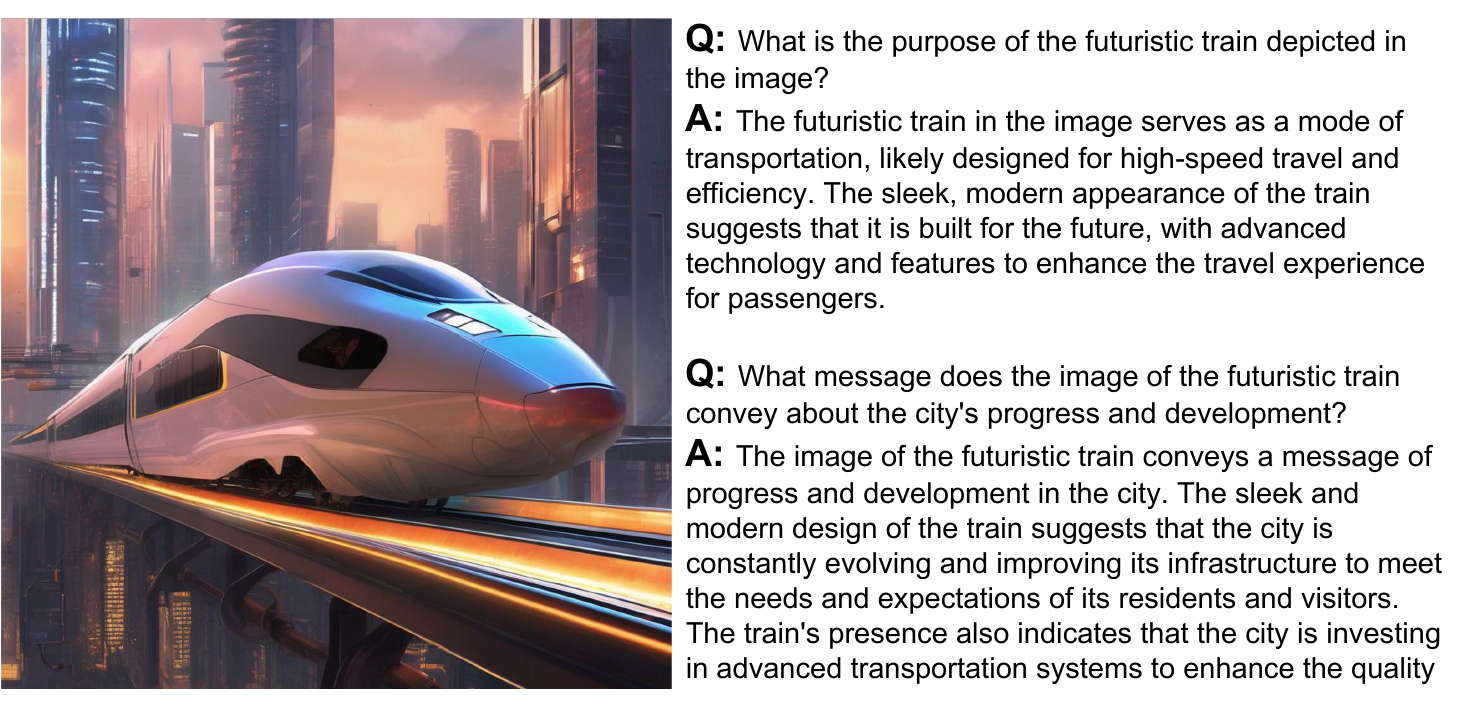}
    \caption{\textbf{Randomly sampled single-image QA generated by our data generation pipeline}.}
    \label{fig:single_qa_exp2}
\end{figure*}

\begin{figure*}[h]
    \centering
    \includegraphics[width=1.0\textwidth]{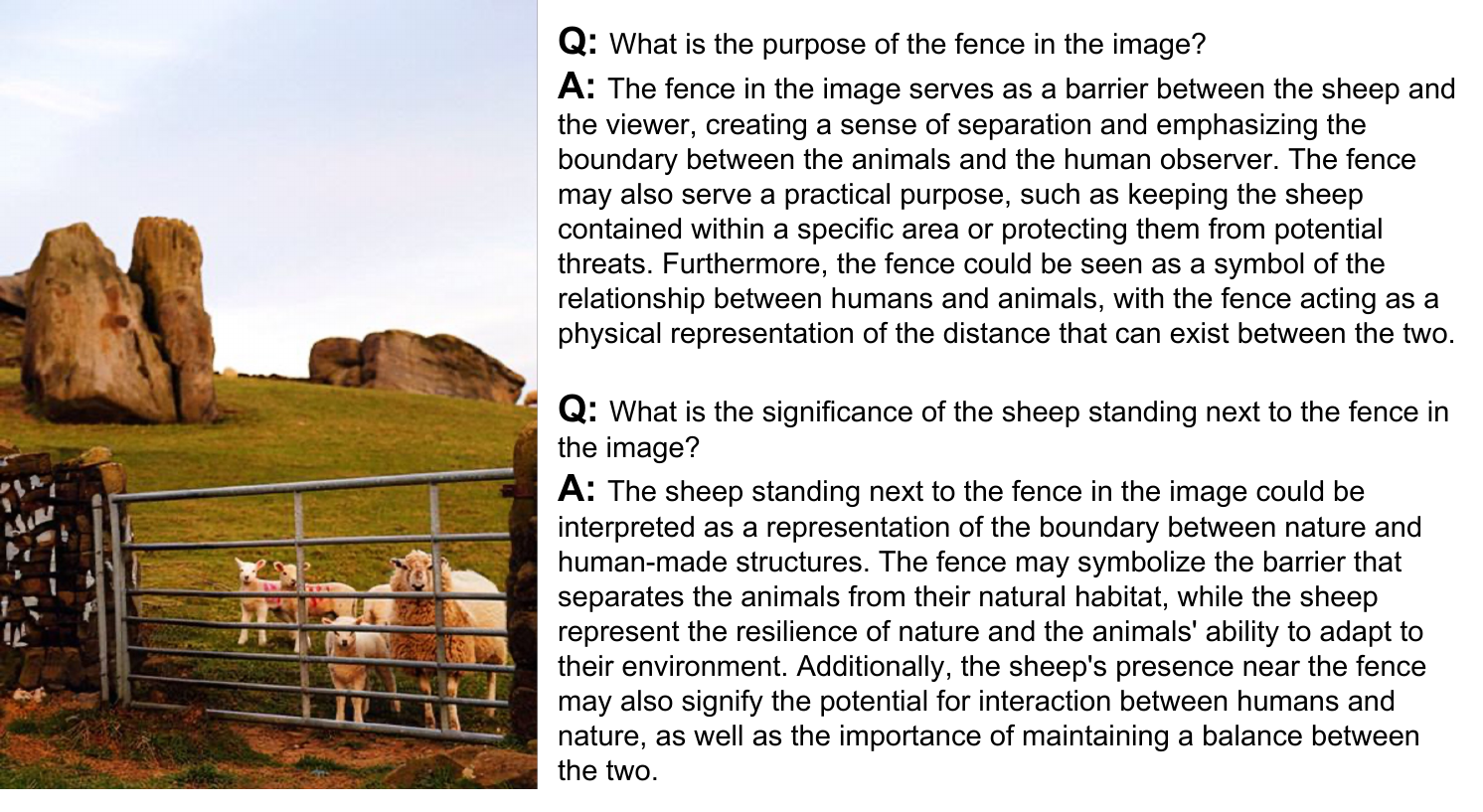}
    \caption{\textbf{Randomly sampled single-image QA generated by our data generation pipeline}.}
    \label{fig:single_qa_exp3}
\end{figure*}

\begin{figure*}[h]
    \centering
    \includegraphics[width=1.0\textwidth]{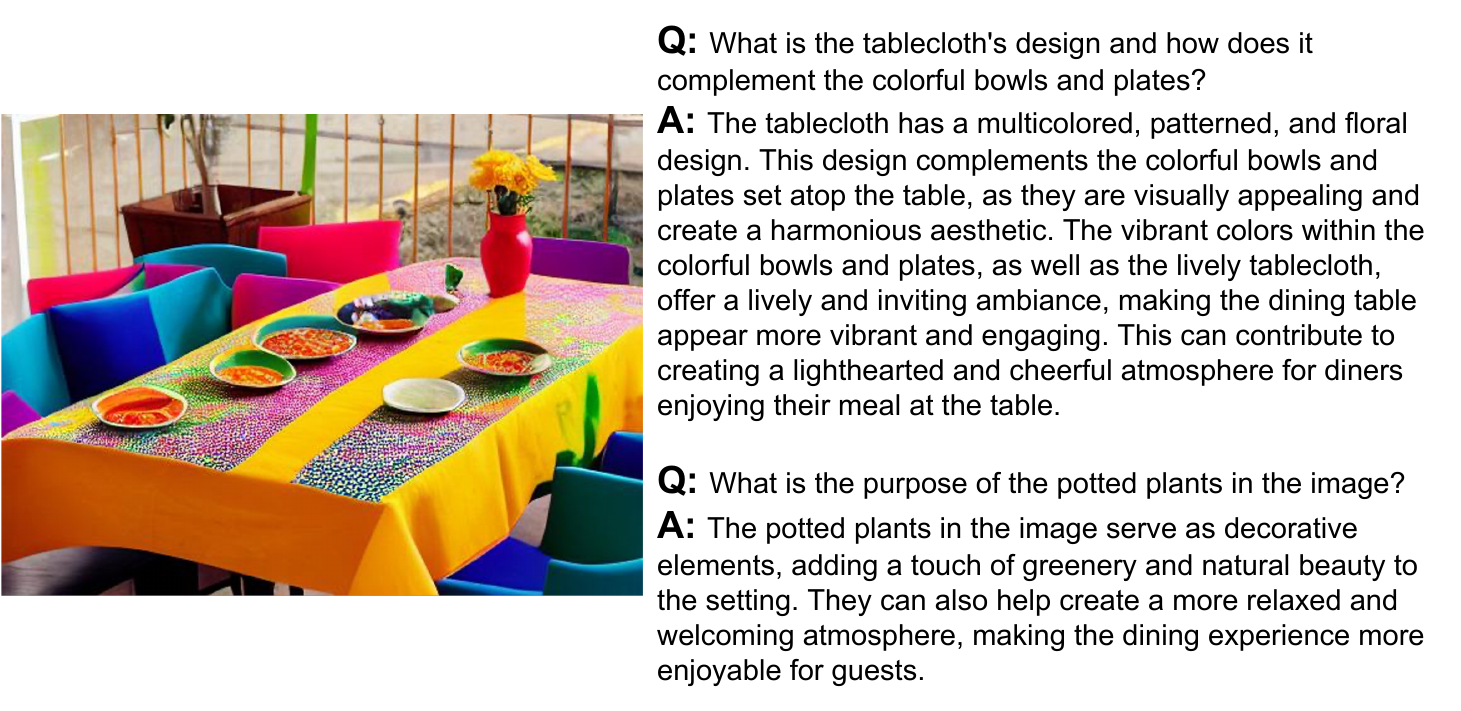}
    \caption{\textbf{Randomly sampled single-image QA generated by our data generation pipeline}.}
    \label{fig:single_qa_exp4}
\end{figure*}

\begin{figure*}[h]
    \centering
    \includegraphics[width=1.0\textwidth]{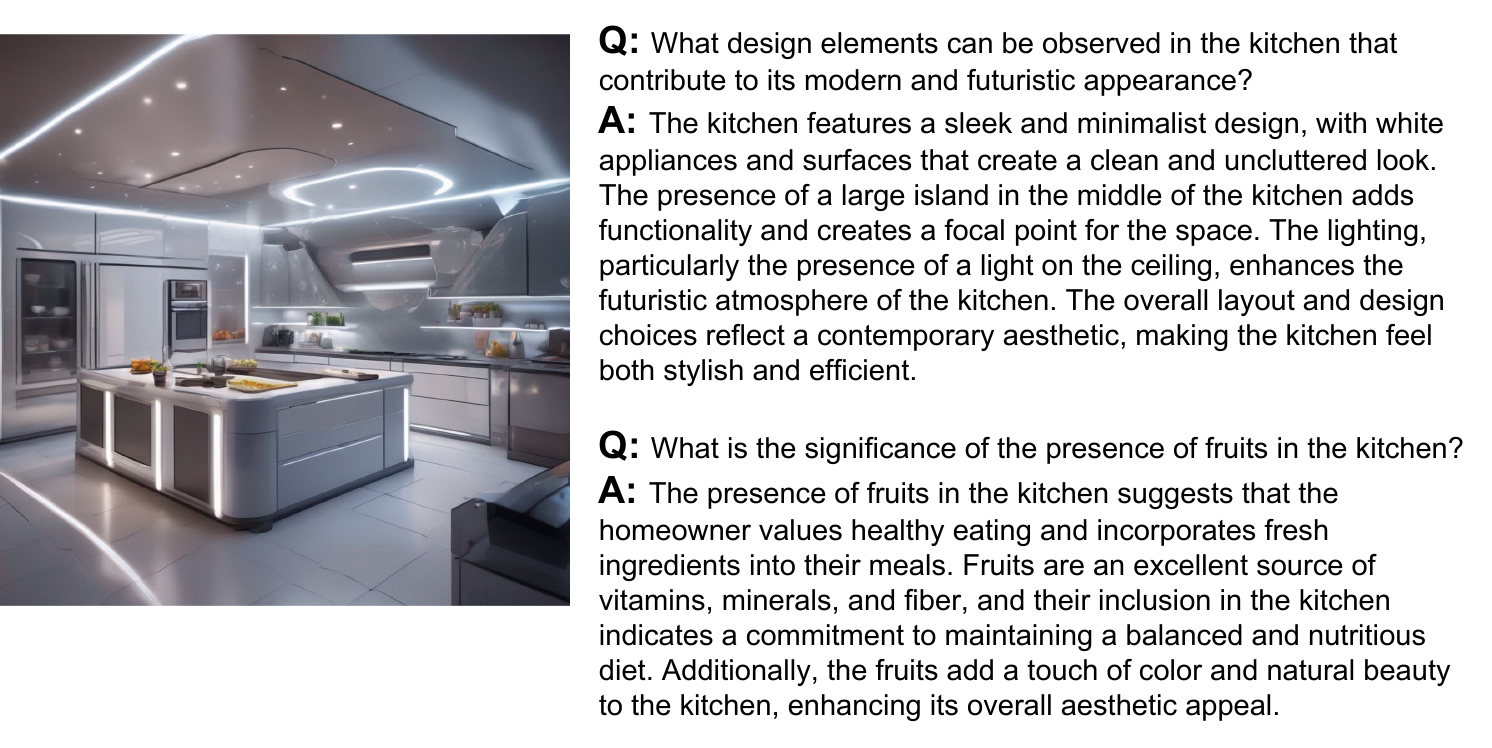}
    \caption{\textbf{Randomly sampled single-image QA generated by our data generation pipeline}.}
    \label{fig:single_qa_exp5}
\end{figure*}

\begin{figure*}[h]
    \centering
    \includegraphics[width=1.0\textwidth]{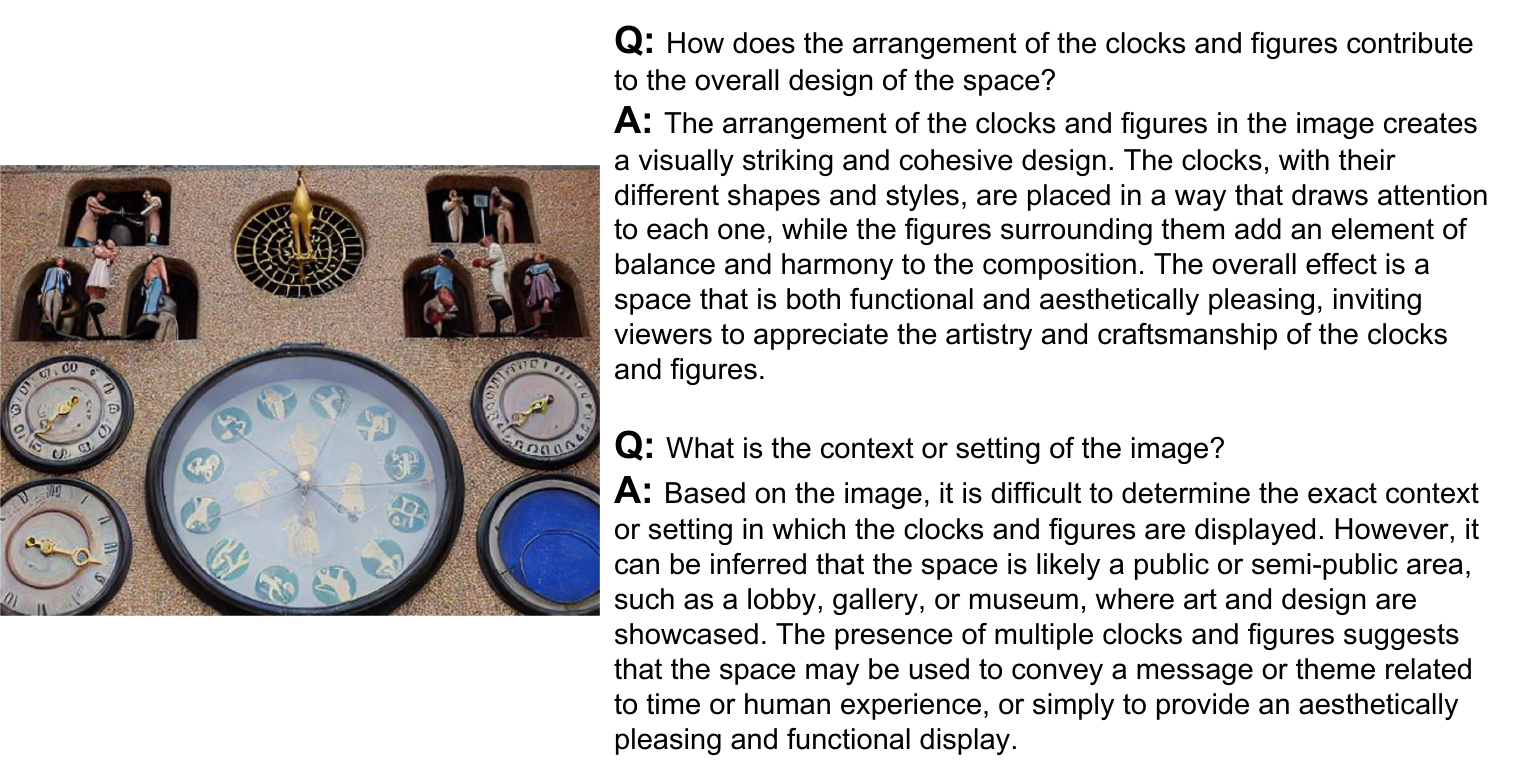}
    \caption{\textbf{Randomly sampled single-image QA generated by our data generation pipeline}.}
    \label{fig:single_qa_exp6}
\end{figure*}

\begin{figure*}[h]
    \centering
    \includegraphics[width=1.0\textwidth]{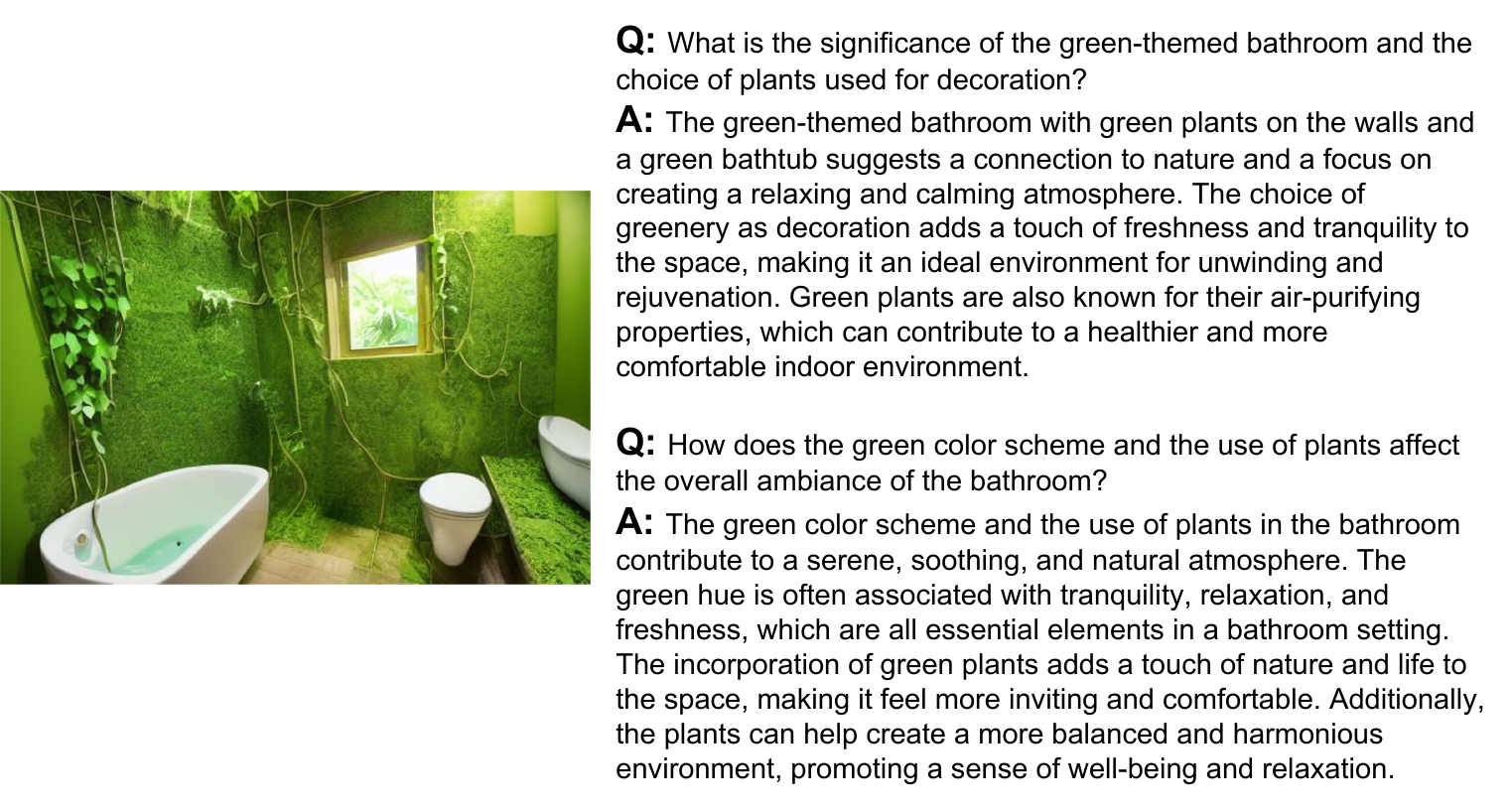}
    \caption{\textbf{Randomly sampled single-image QA generated by our data generation pipeline}.}
    \label{fig:single_qa_exp7}
\end{figure*}

\begin{figure*}[h]
    \centering
    \includegraphics[width=1.0\textwidth]{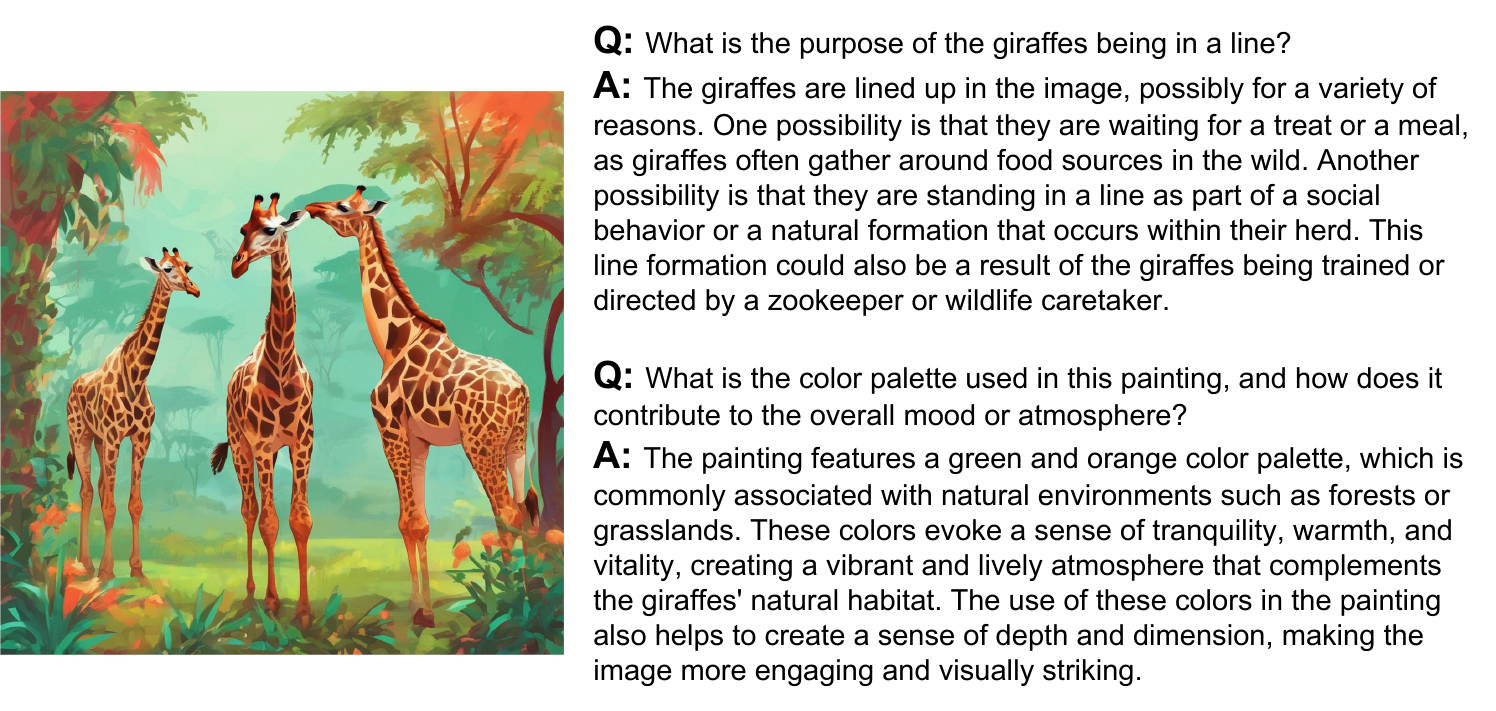}
    \caption{\textbf{Randomly sampled single-image QA generated by our data generation pipeline}.}
    \label{fig:single_qa_exp8}
\end{figure*}

\begin{figure*}[h]
    \centering
    \includegraphics[width=1.0\textwidth]{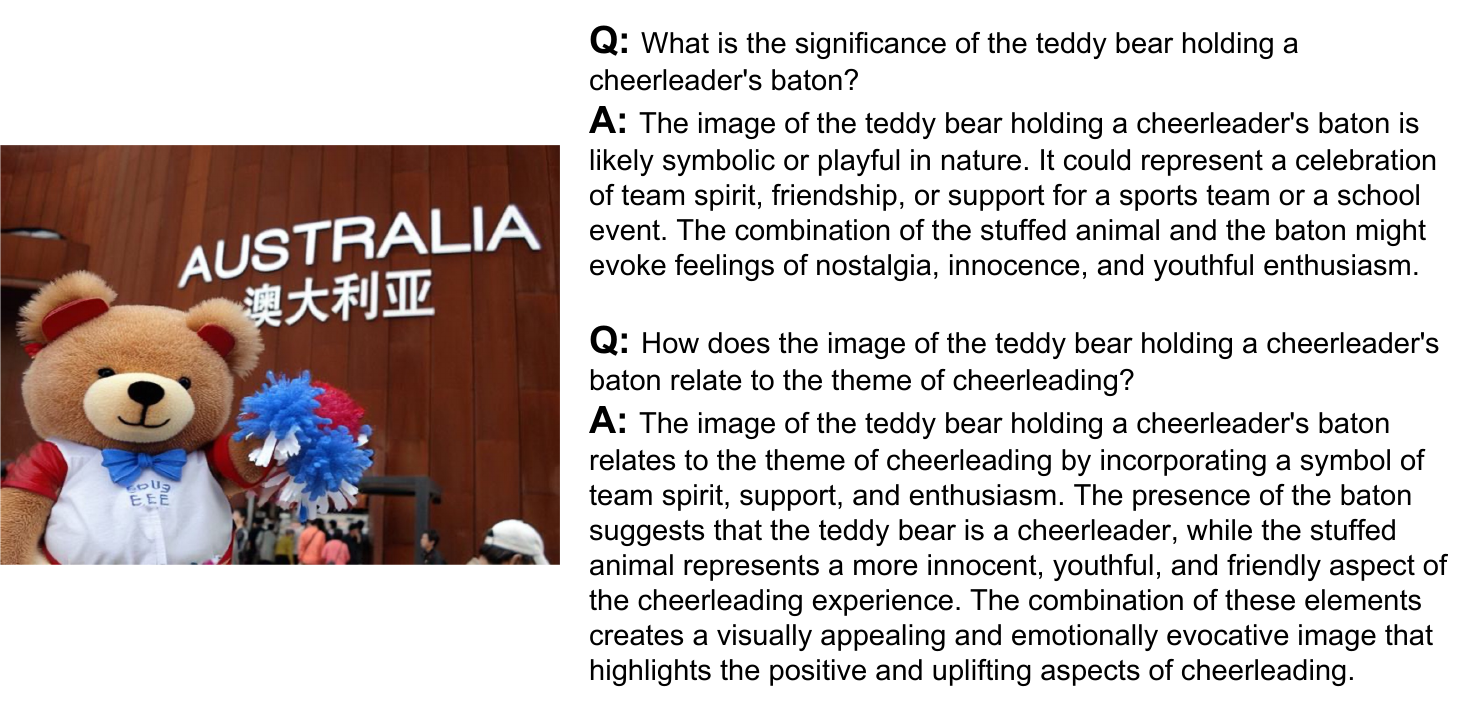}
    \caption{\textbf{Randomly sampled single-image QA generated by our data generation pipeline}.}
    \label{fig:single_qa_exp9}
\end{figure*}

\begin{figure*}[h]
    \centering
    \includegraphics[width=1.0\textwidth]{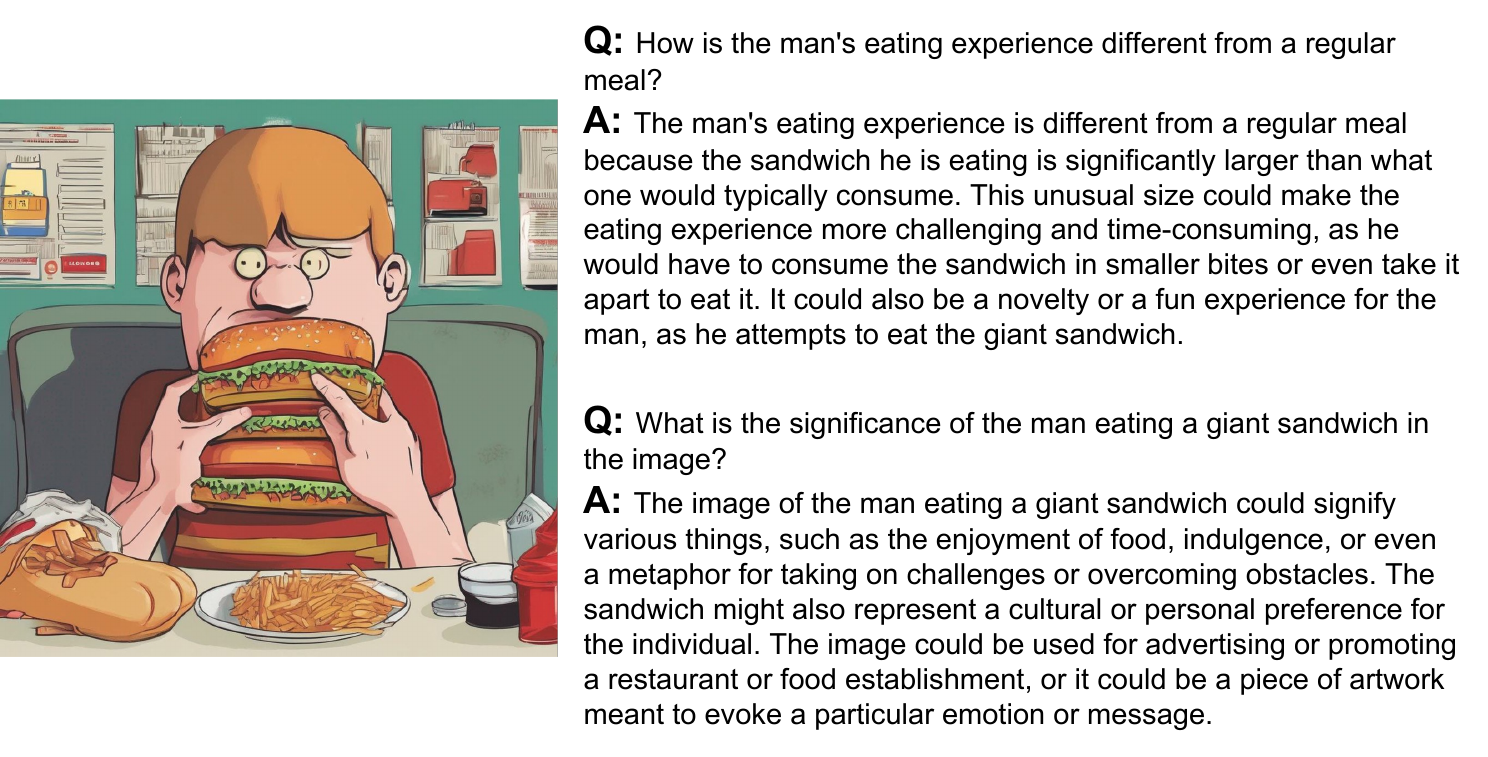}
    \caption{\textbf{Randomly sampled single-image QA generated by our data generation pipeline}.}
    \label{fig:single_qa_exp10}
\end{figure*}

\begin{figure*}[h]
    \centering
    \includegraphics[width=1.0\textwidth]{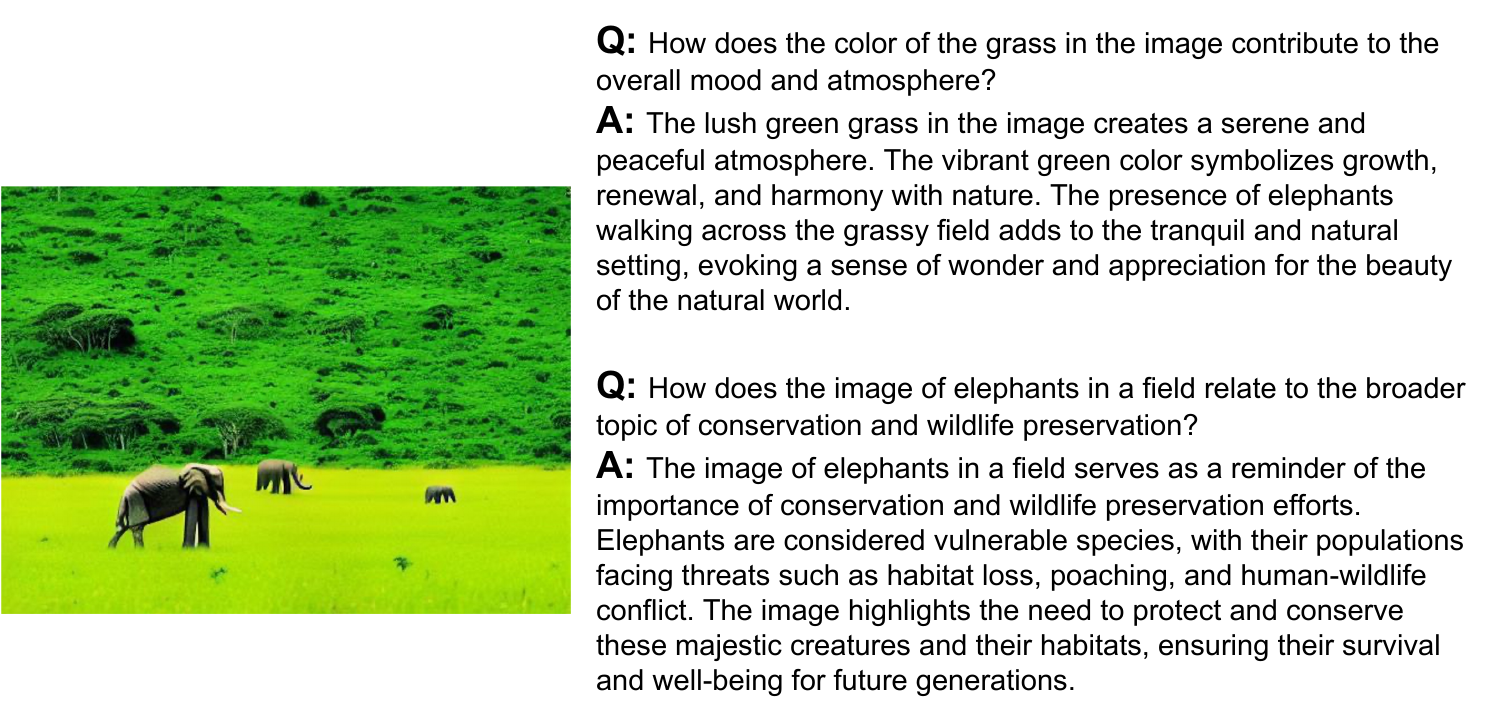}
    \caption{\textbf{Randomly sampled single-image QA generated by our data generation pipeline}.}
    \label{fig:single_qa_exp11}
\end{figure*}

\begin{figure*}[h]
    \centering
    \includegraphics[width=1.0\textwidth]{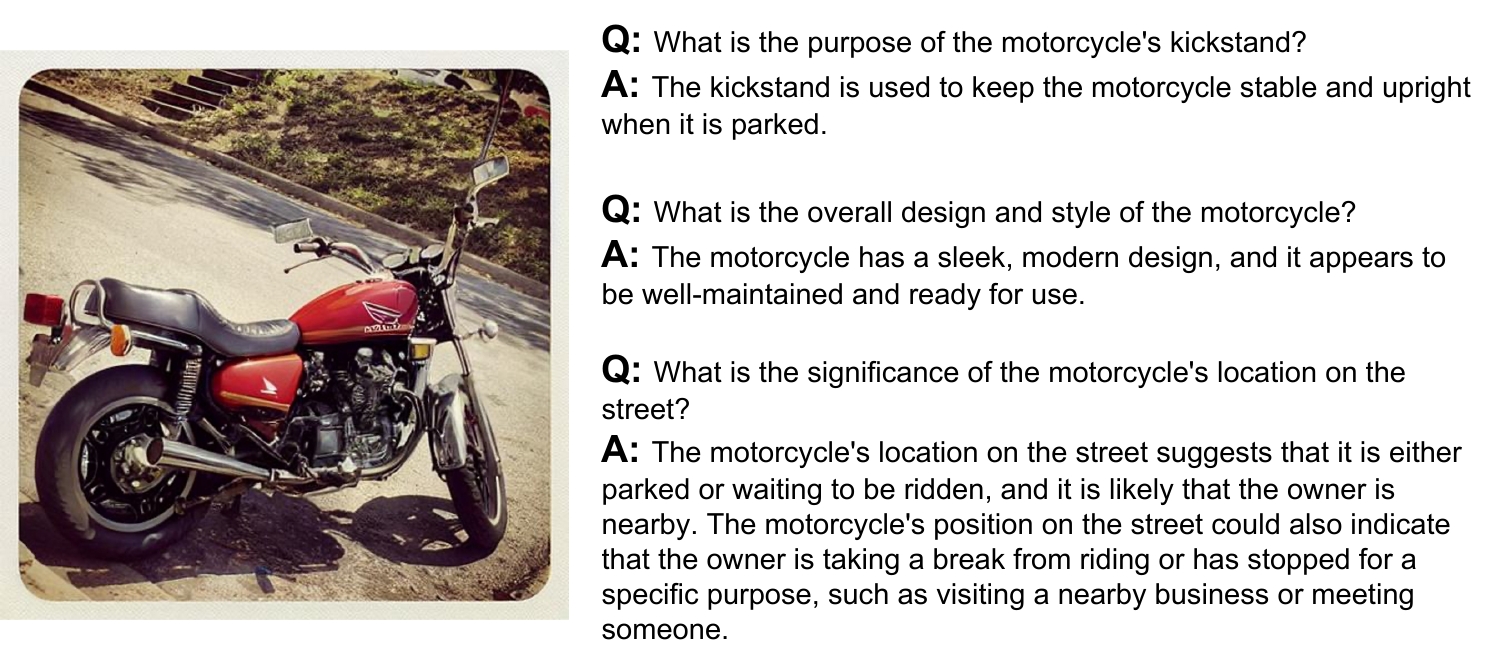}
    \caption{\textbf{Randomly sampled single-image QA generated by our data generation pipeline}.}
    \label{fig:single_qa_exp12}
\end{figure*}

\begin{figure*}[h]
    \centering
    \includegraphics[width=1.0\textwidth]{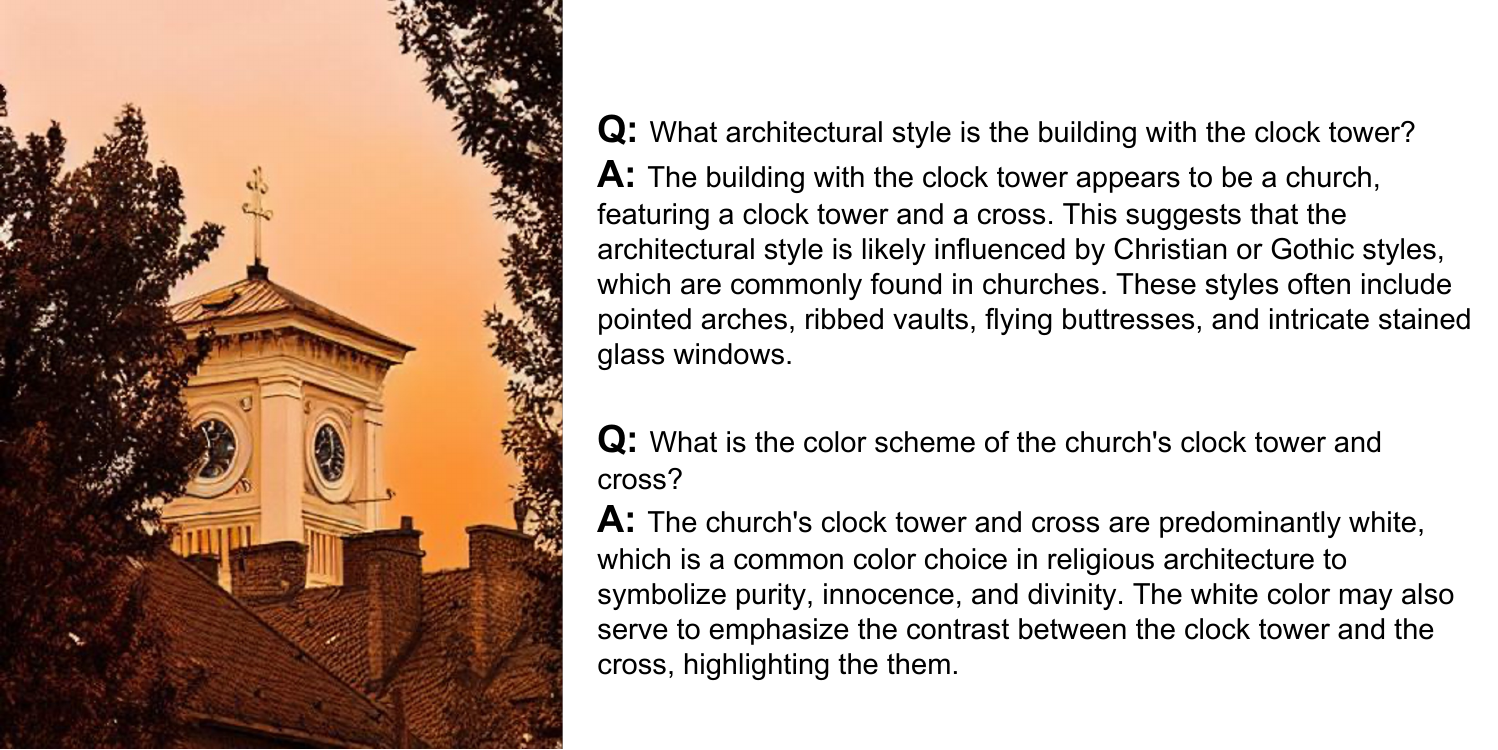}
    \caption{\textbf{Randomly sampled single-image QA generated by our data generation pipeline}.}
    \label{fig:single_qa_exp13}
\end{figure*}

\begin{figure*}[h]
    \centering
    \includegraphics[width=1.0\textwidth]{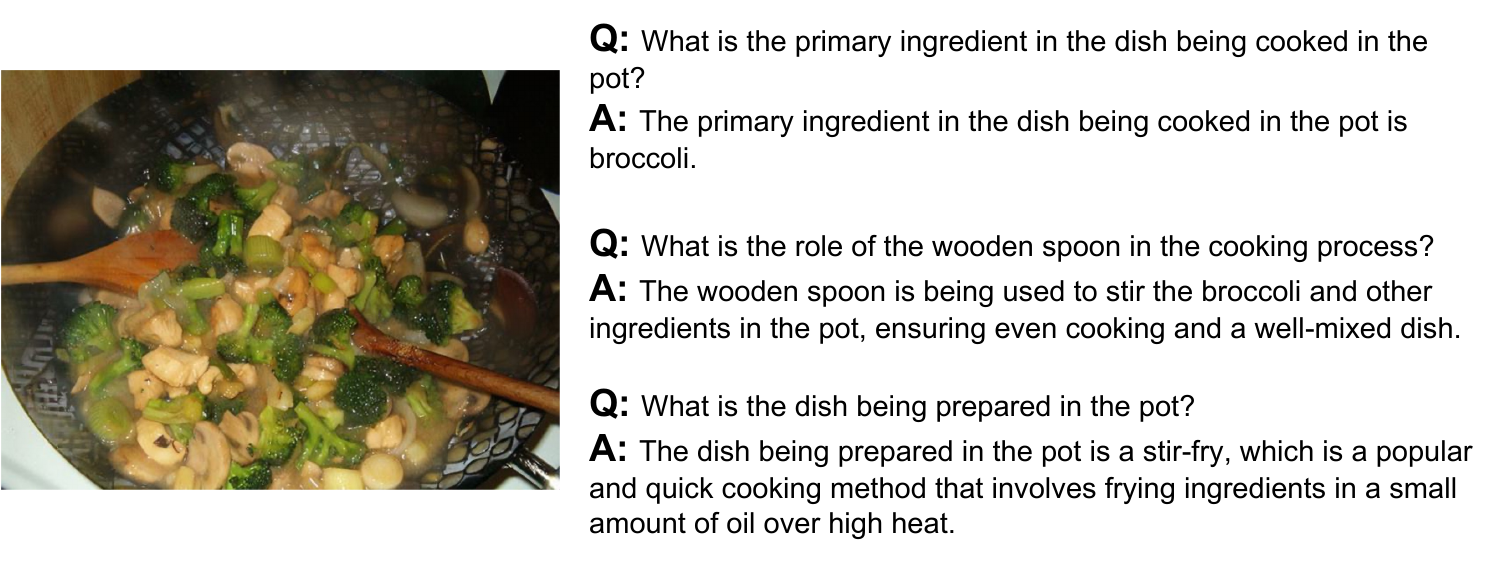}
    \caption{\textbf{Randomly sampled single-image QA generated by our data generation pipeline}.}
    \label{fig:single_qa_exp14}
\end{figure*}

\begin{figure*}[h]
    \centering
    \includegraphics[width=1.0\textwidth]{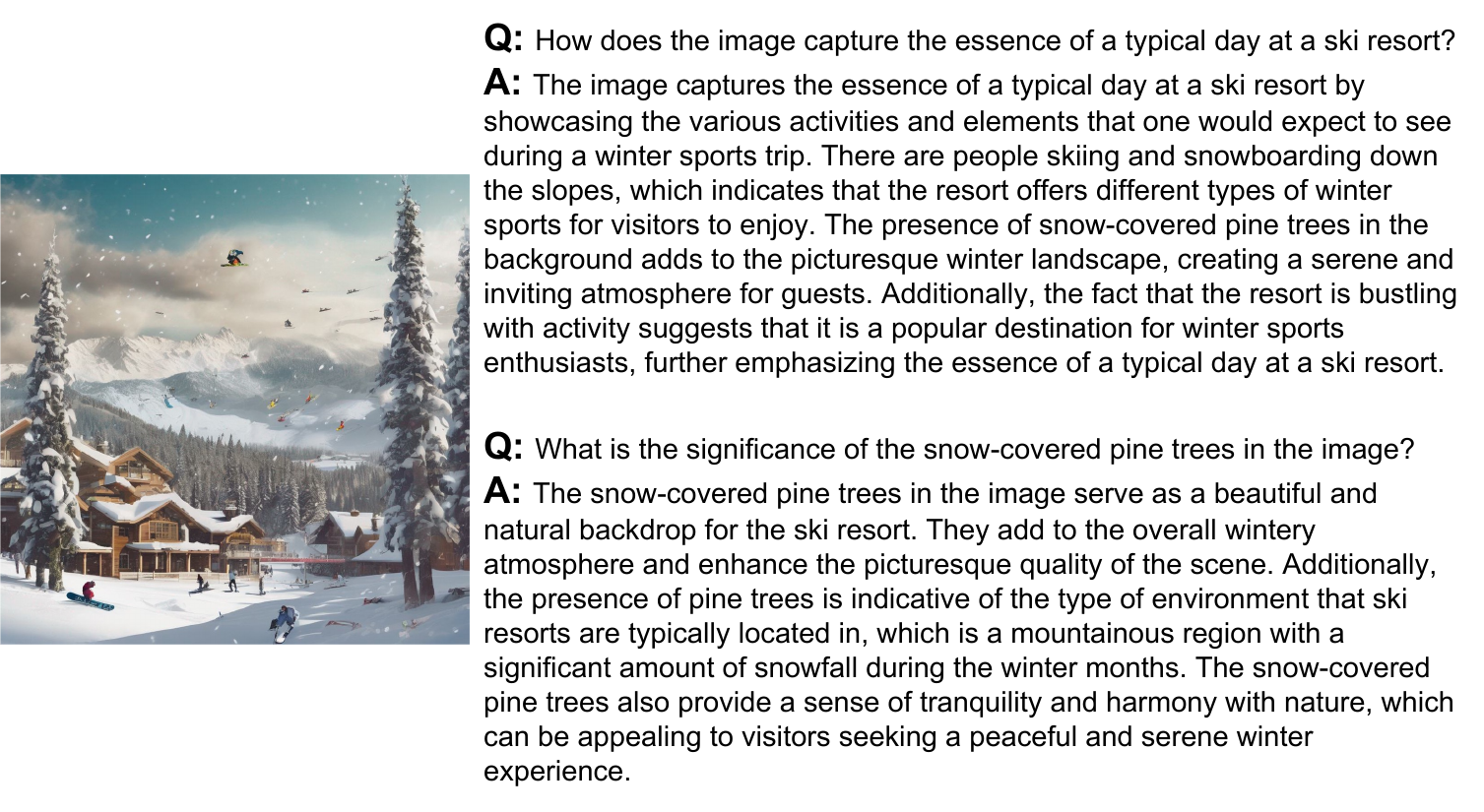}
    \caption{\textbf{Randomly sampled single-image QA generated by our data generation pipeline}.}
    \label{fig:single_qa_exp15}
\end{figure*}